\documentclass[10pt,journal,compsoc]{IEEEtran}
\usepackage{cite}
\usepackage{url}
\usepackage{ragged2e}
\usepackage{epsfig}
\usepackage{graphicx}
\usepackage{amsmath,amssymb} 
\usepackage{algorithm}
\usepackage{algorithmicx}
\usepackage{algpseudocode}
\usepackage{graphics}
\usepackage{threeparttable}
\usepackage{color}
\usepackage[normalem]{ulem}
\usepackage{multirow}
\usepackage{float}
\usepackage{subfig}
\usepackage{amsfonts}
\usepackage{bm}
\usepackage{array}
\usepackage[table]{xcolor}
\usepackage{colortbl}
\usepackage{pifont}
\usepackage{diagbox}
\usepackage{rotating}
\usepackage{booktabs}
\usepackage{overpic}
\usepackage{textcomp}
\usepackage{contour}
\usepackage{enumitem}
\usepackage{makecell}
\usepackage{setspace}
\usepackage{stfloats}
\usepackage[pagebackref=false,breaklinks=false,colorlinks=false]{hyperref}
\usepackage{caption} 
\usepackage{tcolorbox} 
\usepackage[edges]{forest}

\usepackage[switch]{lineno}

\RequirePackage{silence}
\definecolor{mygray}{gray}{.92}

\newcommand{\tabref}[1]{Tab. \ref{#1}}

\newcommand{\figref}[1]{Fig. \ref{#1}}

\newcommand{\secref}[1]{$\S~$\ref{#1}}

\def\ie{\emph{i.e.}}

\def\etc{\emph{etc}}
\def\etal{{\em et al.~}}

\graphicspath{{./}}
\DeclareGraphicsExtensions{.jpg,.pdf,.png}

\def\data{\texttt{CvpINST}}
\def\agent{\texttt{CvpAgent}}

\usepackage{fancyhdr}

\definecolor{fancycolor}{RGB}{64,140,177} 
\let\HeadRule\headrule
\renewcommand\headrule{\color{fancycolor}\HeadRule}

\definecolor{headerbg}{RGB}{252,244,228}  
\definecolor{tablebg}{RGB}{252,244,228} 
\definecolor{linecolor}{RGB}{226,205,159}     
\definecolor{captionbg}{RGB}{251,235,211} 
\definecolor{captiontext}{RGB}{0,0,0} 
\definecolor{myblue}{RGB}{200,220,240}
\definecolor{mygreen}{RGB}{220,240,220}

\begin{document}


\title{Deep Learning in Concealed Dense Prediction} 
\author{Pancheng Zhao$^1$ \footnotemark, Deng-Ping Fan$^1$, Shupeng Cheng$^2$, Salman Khan$^3$, \\Fahad Shahbaz Khan$^3$, David Clifton$^4$, Peng Xu$^2$, and Jufeng Yang$^{156}$
}

\IEEEtitleabstractindextext{
\begin{abstract}
    \justifying
    Deep learning is developing rapidly and handling common computer vision tasks well. It is time to pay attention to more complex vision tasks, as model size, knowledge, and reasoning capabilities continue to improve.
    In this paper, we introduce and review a family of complex tasks, termed Concealed Dense Prediction (CDP), which has great value in agriculture, industry, \etc.
    CDP’s intrinsic trait is that the targets are concealed in their surroundings, thus fully perceiving them requires fine-grained representations, prior knowledge, auxiliary reasoning, \etc.
    The contributions of this review are three-fold:
    (i) We introduce the scope, characteristics, and challenges specific to CDP tasks and emphasize their essential differences from generic vision tasks.
    (ii) We develop a taxonomy based on concealment counteracting to summarize deep learning efforts in CDP through experiments on three tasks. We compare 25 state-of-the-art methods across 12 widely used concealed datasets. 
    (iii) We discuss the potential applications of CDP in the large model era and summarize 6 potential research directions.
    We offer perspectives for the future development of CDP by constructing a large-scale multimodal instruction fine-tuning dataset, \data, and a concealed visual perception agent, \agent.
    Our traceable resource links are \url{https://github.com/PanchengZhao/Concealed-Dense-Prediction}.
\end{abstract}
}
\maketitle
\begin{figure}[!t]
    \centering
    \begin{overpic}[width=\linewidth]{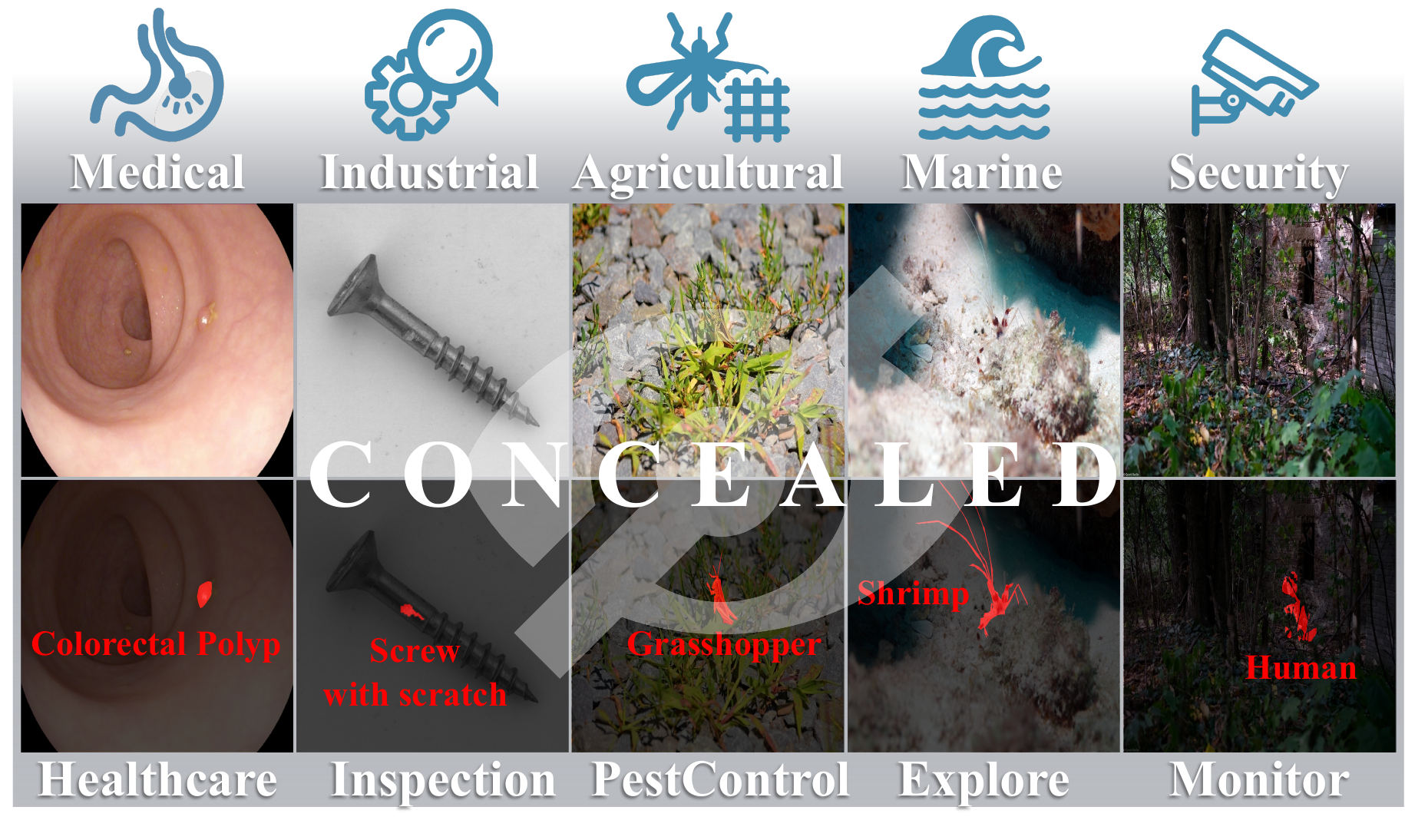}
    \end{overpic}
    \vspace{-20pt}
    \caption{\small Representative concealed scenarios. 
    Targets are concealed in their surroundings, so they may remain unnoticed even when directly observed.
        } 
    \label{fig:concealed_scenarios}
\end{figure}

\IEEEdisplaynontitleabstractindextext
\IEEEpeerreviewmaketitle
\IEEEraisesectionheading{\section{Introduction}\label{sec:}}
%
%
\thispagestyle{fancy}
\renewcommand{\thefootnote}{}  
\renewcommand{\footnotemark}{}  
\setlength{\skip\footins}{0.1in}
\footnotetext{$^1$College of Computer Science, Nankai University, Tianjin, China.
$^2$Department of Electronic Engineering, Tsinghua University, Beijing, China.
$^3$Mohammed Bin Zayed University of Artificial Intelligence, Masdar City, Abu Dhabi.
$^4$Department of Engineering Science, University of Oxford, Oxford, United Kingdom.
$^5$Nankai International Advanced Research Institute (SHENZHEN FUTIAN), Shenzhen, China.
$^6$Pengcheng Laboratory, Shenzhen, China.
E-mail: peng\_xu@tsinghua.edu.cn, yangjufeng@nankai.edu.cn.
}
\IEEEPARstart{T}{raditional} 
computer vision tasks, such as recognition, detection, and segmentation, generally target common objects whose almost fully exposed features make them explicit for computer vision perception, leading to better performance.
%
However, as shown in \figref{fig:concealed_scenarios}, some targets are perfectly concealed in their surroundings. Concealed vision, which aims to perceive these targets, is of great value for agriculture, industry, \etc. 
In recent years, camouflage-themed research has been a preliminary exploration of concealed vision, and the breakthrough of deep learning technology and the emergence of large models \cite{steyvers2025large,schulze2025visual} have injected new vitality into this community. 
%
By adopting new model architectures \cite{chen2024camodiffusion} or introducing pre-trained large models \cite{hu2024relax}, computers have gained advances in their ability to perceive concealed targets, triggering a new wave of research.
Researchers are no longer satisfied with repetitively studying simple traditional tasks, gradually transitioning their attention from explicit vision to the more challenging concealed vision.
As one of the fundamental problems in computer vision, dense prediction focuses on mapping an input image to a complex output structure.
This requires the model to preserve the image's details and comprehensively understand each pixel, making it the best practice for thoroughly mining images for concealed targets.
%
In this context, we introduce the concept of Concealed Dense Prediction (CDP) to unify recent concealment-themed research efforts, marking a preliminary yet meaningful exploration into the concealed visual perception area.

CDP differs from traditional computer vision tasks
\cite{xu2021multigraph,zou2023object} 
by shifting attention from explicit to concealed targets that can evade or deceive vision perception.
This perspective, which classifies computer vision tasks based on the visibility and recognizability of the targets to computer perception, 
rather than the specific implementation methods, offers a unique and valuable framework for understanding the challenges and opportunities associated with CDP. 
Within this framework, each concealment mechanism typically necessitates a specific type of deep model to counteract it, thereby enhancing the visibility of the corresponding concealed target to computer vision perception. 
This underscores the significance of comprehending the specific characteristics of the target and customizing the model architecture based on the concealment mechanism.

Despite its challenges, CDP offers significant application potential across various domains, attracting many researchers to this field. 
%
Consequently, it has been extensively studied and applied in fields such as 
pattern recognition, agriculture (yield estimation, pest detection, and climate inference), rare species protection, medical health (intestinal polyp detection and liver tumor segmentation), artificial intelligence and automation.
This review aims to introduce CDP, provide an outlook on concealed vision, and review related deep learning tasks, to inform researchers and foster future advancements in this field.

\noindent\textbf{$\bullet$ Previous Reviews and Scope.}
There are currently only a limited number of comprehensive reviews and benchmarks to support CDP's further development.
Certain reviews explore the generation of concealment at a theoretical level \cite{stevens2009animalcam,niu2018plant}, while others concentrate on the phenomenon of concealment in specific application scenarios
\cite{shimoni2019hypersectral,fan2021transparent}.
Although some benchmarks aim to summarize tasks and methods for addressing camouflage, they tend to have a narrow focus. Some only include non-deep learning techniques \cite{kulchandani2015moving,mondal2020camouflaged}, some lack discussion of downstream tasks \cite{mondal2022camouflage}, and others only consider image-level tasks \cite{bi2021rethinking,caijuan2022survey,liang2023systematic}.
Until recently, Fan~\etal\cite{fan2023advances} provided a more comprehensive review and several benchmarks for concealed scene understanding (CSU). 
This is the first time camouflage has been categorized as concealment, shifting the focus from the animal itself to the relationship between the scene and the targets.
%
%
Xiao~\etal\cite{xiao2024survey} further reviewed existing tasks in the CSU, providing an extensive benchmark along with quantitative and qualitative analyses of commonly used COD methods, datasets, and evaluation metrics.
Taking inspiration from them, we further expand the definition to present a review of the CDP literature. 
Based on the characteristics of concealment, we establish the connection between \textbf{concealed genesis}, \textbf{deep learning techniques to counteract concealment}, and \textbf{downstream applications}, with a broader scope.

\noindent\textbf{$\bullet$ Contributions.}
We review the state of deep learning development and applications in the CDP area.
In particular:
(i) Starting from the causes of concealment, we summarized CDP tasks in \textbf{14} categories, highlighting their characteristics, challenges, and essential differences from general tasks.
(ii) We develop a detailed taxonomy to cover more than 100 relevant studies under 22 concealment counteracting methods. 
We quantitatively and qualitatively evaluate the performance of state-of-the-art CDP methods on 12 commonly used concealed datasets through theoretical analyses and experimental demonstrations across 3 common dense prediction tasks. Additionally, we explore the generalization performance of methods designed for different concealment mechanisms in other concealed scenarios, aiming to provide insights for developing a unified approach to CDP.
(iii) We summarize the prospects for CDP applications in 4 domains, as well as 6 potential future research directions.
We offer perspectives for the future development of CPD by constructing a large-scale multimodal instruction fine-tuning dataset, \data, and a unified concealed visual perception agent \agent.

\section{Background}

\subsection{Domain-Unique Challenges and Taxonomy}\label{sec:cmou_taxonomy}
As David Marr \cite{noe2002vision} says, ``Vision is the process of discovering from images what  is  present  in  the  world,  and  where  it  is.''
%
%
However, visual information alone is insufficient for complete understanding, as the brain’s inductive reasoning abstracts perception.
%
%
Modern computer vision techniques have been developed mainly by imitating biological vision.
%
Despite high-precision visual sensors, processing programs remain susceptible to noise during inductive reasoning, often leading to errors.
When a target is not visible to visual perception due to a specific pattern that deceives the visual system and results in erroneous reasoning, it is known as a concealed target.
CDP aims to recognize these visual targets that can easily evade recognition by computers or humans in both natural and artificial scenes.

\noindent\textbf{$\bullet$ Unique Challenges.}
Distinguished from ordinary visual perception tasks, the unique challenges of CDP can be summarized as follows:
(i) \textbf{Abstraction}: 
Concealment is closely associated with perception and cognition. 
To understand concealed strategies, one must grasp the perception and cognition of both the seeker and hider.
%
Determining whether an object is concealed, or the degree of concealment, relies on the observer's experience, making it challenging to describe and compare quantitatively.
(ii) \textbf{Invisibility}: 
Concealed targets are rendered invisible to computer vision by inducing erroneous reasoning processes. This complicates feature extraction and pattern recognition.
(iii) \textbf{Diversity}: 
Concealed targets vary widely in type and environment, and concealment mechanisms are equally diverse. 
This diversity challenges the ability of models, designed under specific conditions, to generalize across different types of concealed targets.
(iv) \textbf{Data scarcity}: 
The stealthy nature of concealed targets results in a scarcity of available data. Complex environments, occlusion, and seamless embedding further complicate pixel-level labeling, making data acquisition labor-intensive and time-consuming.

\begin{figure*}[t!]
    \centering
    \subfloat{\includegraphics[width=.65\linewidth]{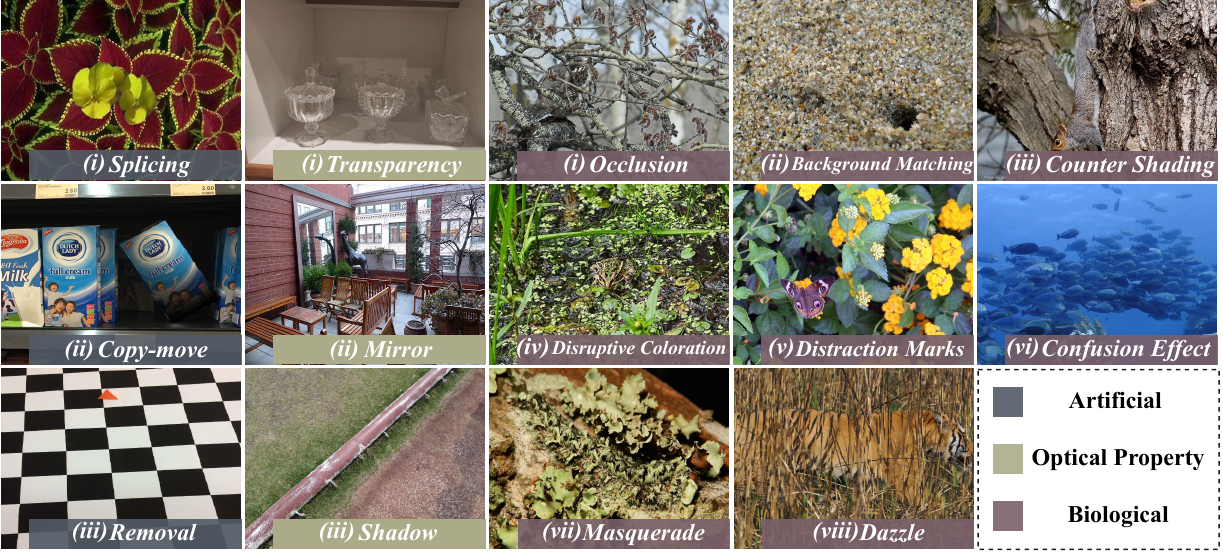}}\hspace{20pt}
    \subfloat{\includegraphics[width=.3\linewidth]{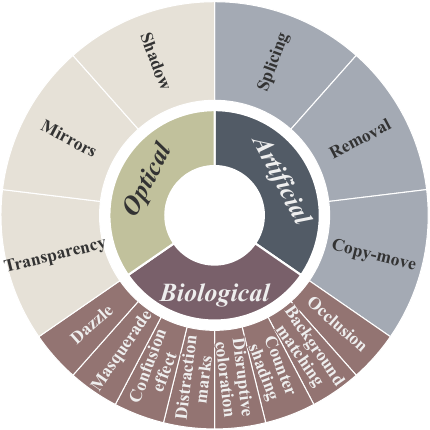}}
    \caption{ The sample gallery and taxonomy of concealment.
    The left shows samples from the different concealment categories in \secref{sec:cmou_taxonomy}. 
    The right shows a hierarchical classification of these concealed categories. 
        } 
    \label{fig:gallery}
\end{figure*}

Existing definitions of concealment are narrow, focusing mainly on camouflaged animals and neglecting other objects with similar properties.
Here we further categorize concealed targets into the following three categories based on their concealment mechanisms: (i) \textbf{biological concealment}, (ii) \textbf{optical property concealment}, and (iii) \textbf{artificial concealment}.
\figref{fig:gallery} visually illustrates our taxonomy and the sample gallery about these three concealment mechanisms.

\subsubsection{Biological Concealment}\label{sec:bio_camo}
To evade predators, many species have evolved natural concealment. Early studies on biological concealment provided strong evidence for Darwin’s and Wallace’s theories of natural selection and adaptation.
%
%
With advances in evolutionary biology, sensory biology, and perceptual and cognitive psychology, the definition of animal concealed patterns has been gradually refined.
%
Researchers have long classified it by appearance, mechanism, and signal-to-noise ratio.
Recent studies integrate cognitive theories and computational neuroscience, identifying six patterns, including transparency and masquerade, while analyzing their formation and limitations \cite{cuthill2019camouflage}.
In this review, we consider biological concealment from the following perspectives:

\textit{(i) Occlusion}:
Birds are naturally inclined to seek shelter in the lush foliage of plants,
 while cheetahs use bushes to conceal themselves and launch sneak attacks. 
When visual detection cannot be avoided, reducing the number of features available for perception is an effective option.
As shown in \figref{fig:gallery}, this can be achieved through occlusion, which involves using other objects in the environment to hide parts of itself and reduce its overall detectability.

\textit{(ii) Background matching}:
Among the various biological concealments, background matching is the most prevalent and extensively studied one.
%
It responds to the interplay between a species' surface color and texture and its habitat, which gradually converge over time under natural selection.
In background matching, the surface of a species exhibits remarkable consistency in terms of color and texture with the background, allowing for seamless embedding.

\textit{(iii) Counter shading}:
It is also known as self-shadowing.
%
Sunlight exposure can cause the sunny side of the target to appear brighter, and this gradient difference in brightness may emphasize its structural information, making it more noticeable to predators.
Special back shading coloring is employed to mitigate the perception of any disparities caused by sunlight, ensuring a consistent appearance.

\textit{(iv) Disruptive coloration}:
%
It refers to a set of special markings on the surface of an object that cause false edges and boundaries through contrast and similarity in color, brightness, and texture.
Due to the natural tendency of human and animal perception to group features, these prominent but noncontiguous visual features significantly reduce their recognition effectiveness. See \figref{fig:gallery} (the second row).

\textit{(v) Distraction marks}:
Some special patches serve as marks to divert the predator’s attention away from accurately identifying the global object, thereby acting as a distraction strategy.
%
This strategy achieves feature concealment by impairing the observer's perception of other nearby stimuli with a stronger stimulus.
%

\textit{(vi) Confusion effect}:
%
Observations of predator-prey interactions reveal that large, synchronized groups reduce an individual’s capture risk. The confusion effect arises as predators struggle to process spatial information from multiple similar targets, with its effectiveness depending on group size and similarity.

\textit{(vii) Masquerade}:
Some organisms naturally resemble inedible objects like leaves or rocks, deceiving predators and avoiding attacks. This differs from crypsis, which blends into the background, and mimicry, which imitates other species for advantage.
\textit{(viii) Dazzle}:
Striped patterns in animals like zebras, fish, and snakes create a dazzling effect in motion, disrupting perception of speed, trajectory, and shape, making tracking difficult. This high-contrast pattern serves as an active concealment strategy, leading to misjudgments.

\subsubsection{Optical Property Concealment}\label{sec:opt_camo}
%
%
When light interacts with an object’s surface, its absorption, reflection, transmission, or refraction creates different visual effects, largely influenced by material properties.
%
Detecting such optical effects is challenging and often leads to misleading results. 
Based on the different effects of light transmission and reflection on object imaging, we classify optical property concealment into the following three categories:

\textit{(i) Transparency}:
%
Transparency occurs when light passes through a medium with minimal reflection, refraction, or absorption. The visual perception of these mediums result from the character of the scene behind them.
%
%
%
%
Transparent objects, like glasses, bottles, and windows, are common in daily life, often serving functional or decorative purposes. 
%
These transparent targets, lacking distinct visual features and blending with their background, present significant challenges for computer vision.

\textit{(ii) Mirrors}:
%
Mirrors are common in daily life, found in places like malls, homes, and restrooms.
Unlike transparent objects, they reflect light creating an upright, equal-sized virtual image.
This unique property of mirrors poses challenges for various computer vision tasks, such as depth estimation, scene reconstruction, and semantic segmentation.
%

\textit{(iii) Shadow}:
When an object blocks light, it casts a shadow, which can offer cues on lighting, object pose, and orientation for 3D scene understanding.
However, shadow detection is challenging due to their variable appearance, similarity to objects or backgrounds, and occlusions.
Reliable detection requires a holistic understanding of lighting, object shapes, and material properties.

\subsubsection{Artificial Concealment}
%
%

Computers receive digital images or videos as visual feature inputs, and such data is highly vulnerable to human tampering.
Image tampering involves the artificial modification or manipulation of an image to deceive viewers or distort information. 
The main types of image tampering include splicing (pasting a target from one image onto another), removal (deleting a target and filling the area with surrounding pixels), and copy-move (copying a region and pasting it elsewhere within the same image).

We categorize these tampering operations as artificial concealment. 
Unlike the previous two types of concealment, artificial concealment does not rely on features resembling the environment or special visual effects. Instead, it depends more on high-level semantic reasoning and smooth articulation at the underlying feature level, making the tampered target appear as if it was originally part of the image.

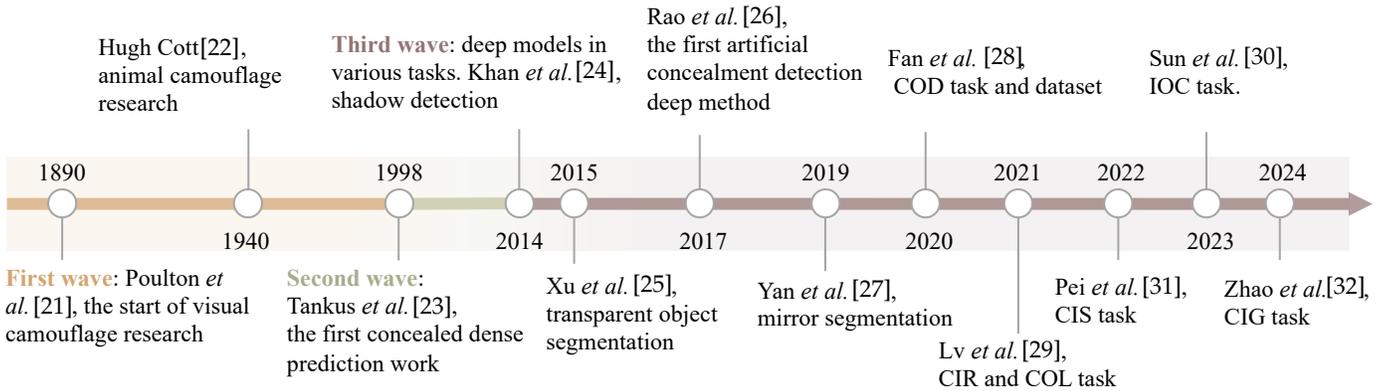
\begin{figure*}[t!]
    \centering
    \begin{overpic}[width=1\linewidth]{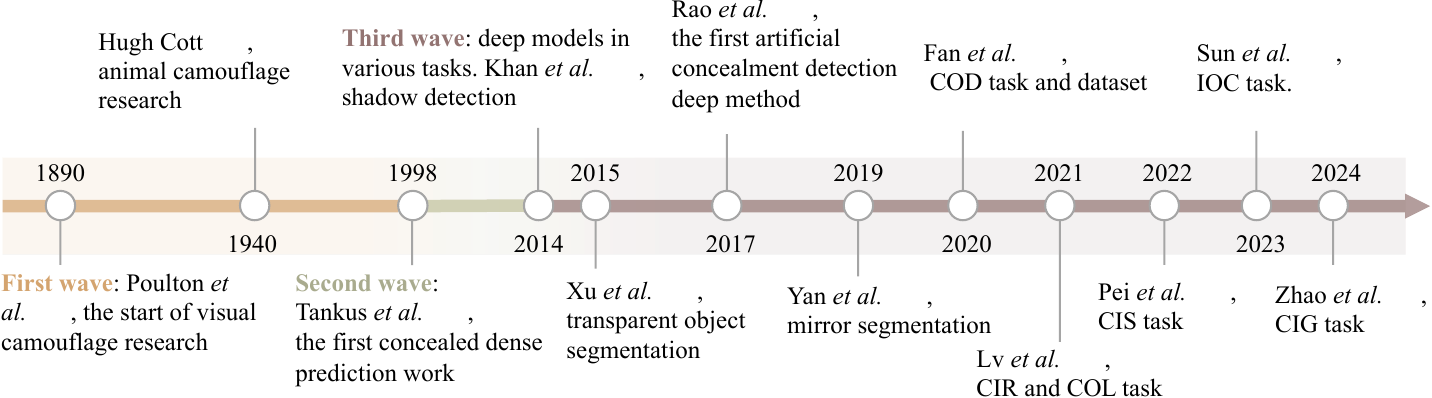}
        \small
        \put(1.6, 5.4){ \cite{poulton1890colours}}
        \put(13.8, 24.3){ \cite{cott1940adaptive}}
        \put(29.3, 5.4){ \cite{tankus1998detection}}
        \put(41.2, 22.6){ \cite{khan2014automatic}}
        \put(45.4, 7){ \cite{xu2015transcut}}
        \put(53.4, 26.6){ \cite{rao2016deep}}
        \put(61.4, 6.6){ \cite{yang2019my}}
        \put(71, 23.6){ \cite{fan2020COD10K}}
        \put(73.8, 2.2){ \cite{lv2021simultaneously}}
        \put(90, 23.6){ \cite{sun2023iocformer}}
        \put(82.6, 6.8){ \cite{pei2022osformer}}
        \put(96, 6.8){ \cite{zhao2024camouflaged}}
    \end{overpic}
    \caption{\small  A simplified chronicle of CDP, 
    including three major waves in the development of CDP and some of the milestones that have advanced the field.
    \textbf{COD:} camouflaged object detection. \textbf{CIR:} concealed instance ranking. \textbf{COL:} concealed object localization. \textbf{COS:} concealed object segmentation. \textbf{IOC:} indiscernible object counting. \textbf{CIG:} camouflaged image generation.
        } 
    \label{fig:history}
\end{figure*}

\subsection{A Brief History of CDP}
The study of concealed vision can be traced back to 1890 \cite{poulton1890colours},
since which researchers have observed that certain visual patterns in animal protective coloration could impede human visual perception.
This observation initiated the first wave of research on concealed vision. 
Subsequent researchers delved into the mechanisms behind concealment, building on these initial ideas and laying the foundation for contemporary concealed vision research.
Following Tankus~\etal~\cite{tankus1998detection}'s first integration of concealed objects into computer vision perception tasks, numerous approaches based on hand-crafted features emerged for detecting concealed objects. 
These methods aimed to distinguish between normal and concealed objects using direct visual features such as color, texture, and optical flow. 
However, their performance and generalization were limited, resulting in restricted adoption.
The advent of deep learning techniques marked the third wave of CDP. 
Leveraging deep neural networks' powerful feature extraction capabilities, models achieved significant improvements in perceiving simple concealed objects. 
During this phase, researchers shifted their focus towards different concealment modalities. They defined various CDP tasks, including transparent object segmentation, concealed object detection, mirror segmentation,  indiscernible object counting, image tampering localization, and more.
\figref{fig:history} illustrates three major waves in the evolution of CDP and highlights several key milestones.

\section{Concealment Counteracting Techniques}
\begin{figure*}[t!]
    \centering
    \begin{overpic}[width=1\linewidth]{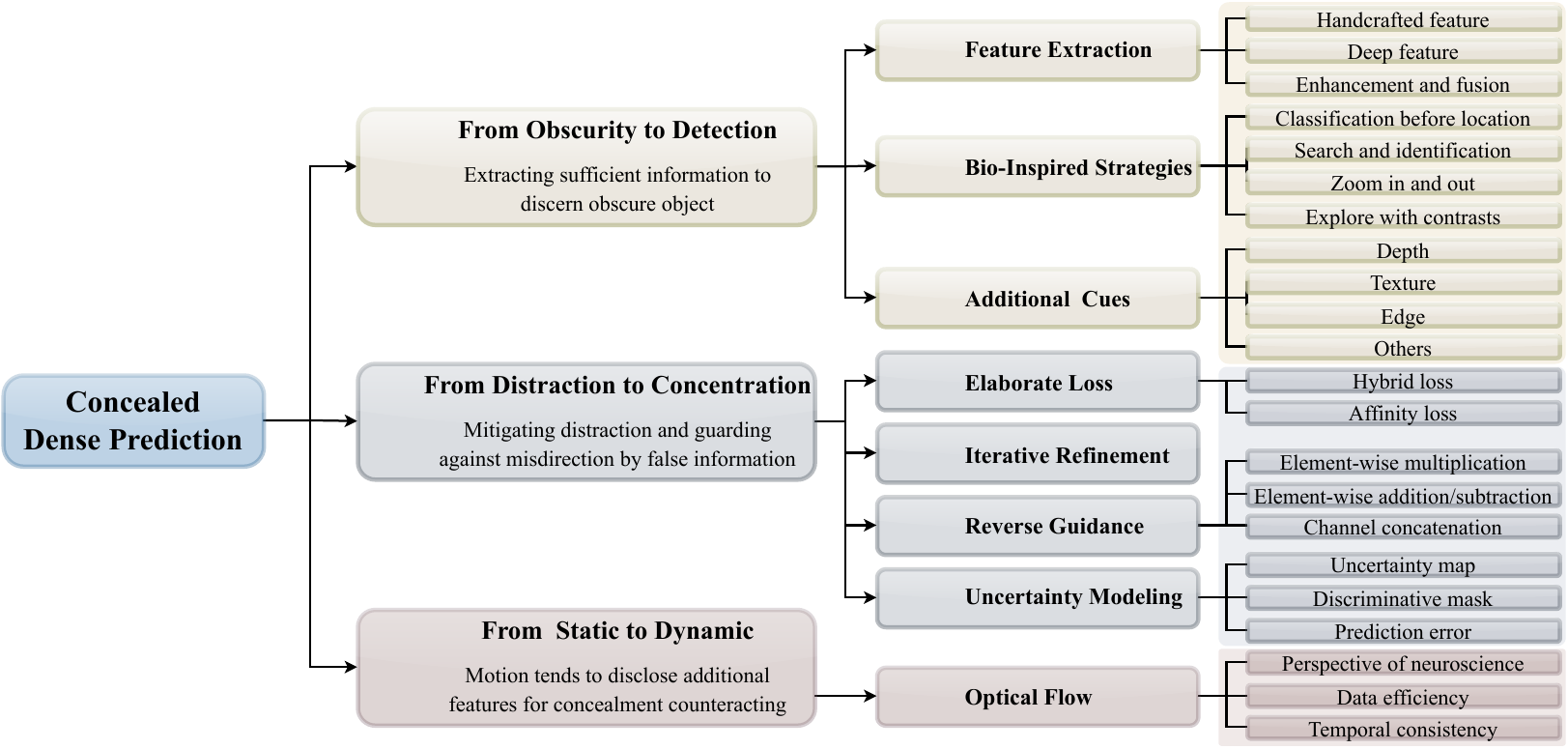}
        \put(22.8, 36.55){ \secref{sec:From_Obscurity_to_Detection}}
        \put(22.8, 20.45){ \secref{sec:From_Distraction_to_Concentration}}
        \put(22.8, 4.6){ \secref{sec:From_Static_to_Dynamic}}

        \put(56.5, 44){\small \secref{sec:Feature_Extraction}}
        \put(56.5, 36.5){\small \secref{sec:Bio-inspired_Strategies}}
        \put(56.5, 28.2){\small \secref{sec:Additional_Cues}}
        \put(56.5, 22.8){\small \secref{sec:Elaborate_Loss}}
        \put(56.5, 18.2){\small \secref{sec:Iterative_Refinement}}
        \put(56.5, 13.7){\small \secref{sec:Reverse_Guidance}}
        \put(56.5, 9.1){\small \secref{sec:Uncertainty_Modeling}}
        \put(56.5, 2.8){\small \secref{sec:From_Static_to_Dynamic}}
    \end{overpic}
    \caption{\small A tree diagram of the CDP method taxonomy.
    CDP methods have been categorized into three classes based on the strategy of concealment, namely extracting detailed features, countering distraction strategies, and mining cues from motion. 
        } 
    \label{fig:method_tax}
\end{figure*}
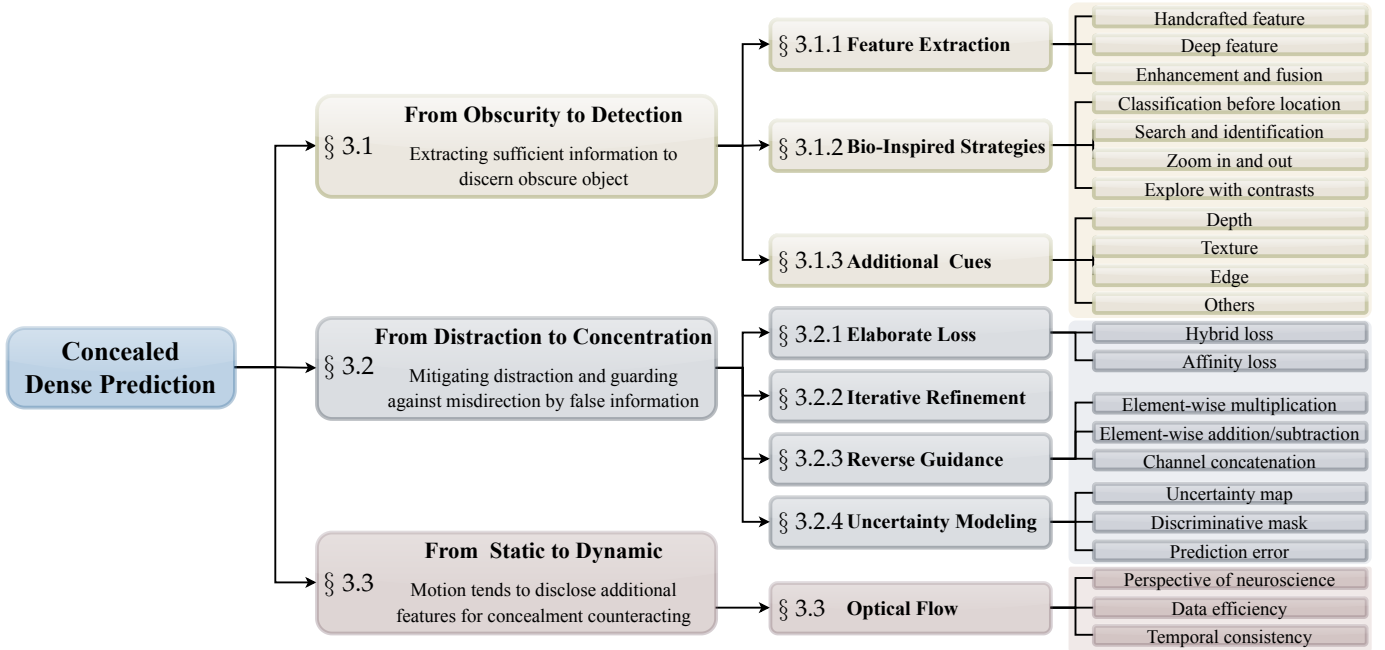

This section aims to provide an overview of deep learning related tasks and methods from the perspective of concealment counteraction.
As shown in \figref{fig:method_tax}, based on how to counteract 
concealment, CDP tasks can be divided into three approaches: searching for more detailed cues to discover objects, preventing false information from causing misdirection, and utilizing additional cues leaked through motion.
The advantages of this task taxonomy are as follows: 
(1) Broad scope: It covers various computer vision tasks, providing an overview of CDP.
(2) Direct and efficient: Tasks are categorized based on concealment counteraction, enabling the transfer of solutions across tasks sharing similar concealment patterns.
\subsection{From Obscurity to Detection}
\label{sec:From_Obscurity_to_Detection}
When viewers encounter a typical and predictable scene, they are less likely to scrutinize it closely.
Consequently, organisms in nature tend to adopt strategies such as protective coloration and mimicry to blend into their environments.
Similarly, objects placed against a matching background with similar surface features also tend to escape notice.
As explained in \secref{sec:bio_camo}, the purpose of concealment is to avoid detection, and this is the most straightforward approach that is widely used across various concealment mechanisms.
%
To counteract this, extracting more effective information from visual features is crucial.
We will delve into three aspects of strategies for extracting effective information to detect concealed targets: feature extraction, bio-inspired strategies, and additional cues.
%


\subsubsection{Feature Extraction}
\label{sec:Feature_Extraction}
Preprocessing visual features is essential for refining and optimizing them, whether the goal is to locate a concealed target or to capture subtle cues from an occluded object.

\noindent\textbf{$\bullet$ Handcrafted feature.}
Traditional approaches heavily rely on handcrafted features tailored for specific tasks to capture intricate details of concealed targets.
%
To visually distinguish concealed targets from their environments, prevalent handcrafted features such as HOG, LTP, and SIFT, are commonly employed to characterize images and reveal subtle distinctions.
However, the applicability of these generic handcrafted features is limited across diverse concealment modalities. The unique characteristics and variations within each concealment modality often render them less effective, necessitating task-specific tailoring.
Li~\etal\cite{li2018wavelet} observed that even in scenarios with high background matching, minute differences between foreground and background can be highlighted in specific wavelet bands, facilitating the detection of concealed targets through frequency features. 
Despite the subtlety of concealment in transparent objects, McHenry~\etal\cite{mchenry2005finding} made a breakthrough by leveraging the systematic distortion in background texture at object boundaries and the strong highlights typical of glass surfaces for detection.
%
In detecting artificial concealment, signal processing methods based on image compression and noise analysis are also employed to locate concealed areas.

\noindent\textbf{$\bullet$ Deep feature.}
Deep Convolutional Neural Networks (CNNs) have unified the methodology for feature extraction.
Their inherent hierarchical and multi-scale structure allows for the extraction of effective feature representations from images.
%
Merely utilizing a pre-trained backbone network is sufficient for acquiring effective representations of the input image across various layers and scales. This streamlined and convenient approach effectively supersedes the need for traditional handcrafted features.
%
The transformer architecture has further enhanced the feature extraction capability of neural networks.
Its application holds significance in tasks related to CDP, such as CamoFormer \cite{yin2022camoformer}, DQNet~\cite{sun2022dqnet}, Trans2Seg \cite{hou2022glass}, and SATNet \cite{huang2023symmetry}, owing to its capacity to capture long-distance connections and preserve spatial information.

\noindent\textbf{$\bullet$ Enhancement and fusion.}
Furthermore, the extracted features are enhanced to obtain more effective representations.
In addition to incorporating commonly used receptive field (RF) modules \cite{liu2018receptive} and feature pyramid networks (FPN) architectures \cite{lin2017feature} to expand the network's receptive field and focus on the multi-scale features of the target, more methods now incorporate feature fusion approaches designed based on concealment mechanisms.
To better detect targets perfectly embedded in the background, employing reasoning with rich contextual information is also a viable approach \cite{mischler2024contextual}.
In this regard, some methods \cite{dong2021accurate,sun2021context,chen2022camouflaged,zhang2022preynet,dong2024camouflaged,hu2024efficient} introduce the interactive attention mechanism to help the model locate and retain challenging parts of the features while considering contextual information. This mechanism aims to eliminate redundancies and achieve an efficient fusion of features.
Additionally, grouping features along the channel dimension facilitates the exploration of interactions between different channels, as demonstrated in many methods \cite{fan2021concealed,pang2022zoom,zhou2022feature}.
Lacking discriminative features, the detection of optical concealment often relies on contextual relationships in the scene.
This leads to models that incorporate parallel feature streams and multi-scale refinement modules. 
Conversely, in artificial concealment, while images appear harmoniously integrated at a perceptual level, inconsistencies in texture and structure at a lower level serve as crucial detection indicators.
These studies motivate researchers to investigate more effective feature extraction strategies to capture subtle differences between concealed targets and backgrounds.

\subsubsection{Bio-Inspired Strategies}
\label{sec:Bio-inspired_Strategies}
Camouflage originates from biological interactions between predators and prey, characterized by a continuous cycle of concealment and detection. This dynamic interplay not only drives the evolution of concealment strategies but also offers valuable insights to counteract concealment.
The visual perception process is time-varying, where the content in a complex scene does not arrive instantaneously but is processed over an extended period, transitioning from the acquisition of simple stimuli to full understanding.
%

\noindent\textbf{$\bullet$ Classification before location.}
Implementing a multitasking framework through a linear joint classification task can emulate how humans acquire prior knowledge during the pre-detection phase \cite{le2019anabranch}.

\noindent\textbf{$\bullet$ Search and identification.}
Similarly, predators initially search for potential objects within a large area and then carefully identify them within regions of interest.
Some CNN architectures mimic this two-stage hunting process by expanding the receptive field and iteratively extracting detailed texture features. These approaches effectively improve detection accuracy \cite{fan2020COD10K,fan2021concealed} and have been widely adopted by researchers \cite{wang2021d,mei2021camouflaged,zhang2022preynet}.

\noindent\textbf{$\bullet$ Zoom in and out.}
%
Organisms maintain a consistent body size to avoid detection, as unusual sizes greatly reduce their ability to stay hidden. 
Very small targets can also be difficult for the visual system to perceive. Predators counteract this by adjusting their distance to alter the perceived size of their prey, disrupting concealment.
Deep learning models\cite{zhu2021global,pang2022zoom,jia2022segment,pang2024zoomnext}  mimic this mechanism by applying zooming in and out operations on visual data to detect and segment camouflaged objects. Zooming in captures local details, while zooming out captures global features, helping the model understand the scene and identify subtle differences.

\begin{figure*}[t]
    \vspace{10pt}
    \centering
    \begin{overpic}[width=.98\linewidth]{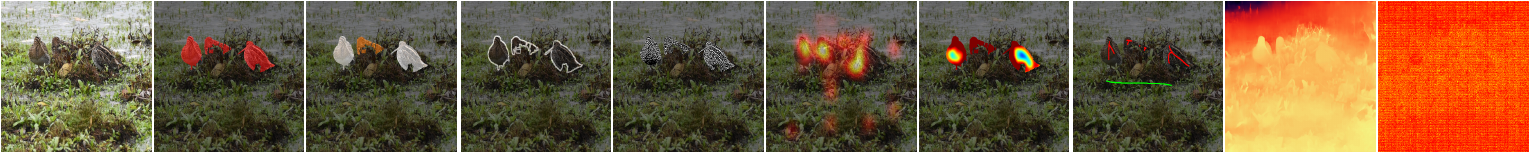}
    \end{overpic}
    \caption{\small Samples of annotation and additional cue used in CDP method.
    Left to right: original image, object annotation \cite{fan2020COD10K}, rank annotation \cite{lv2021simultaneously}, edge annotation \cite{fan2020COD10K}, texture annotations \cite{zhu2021inferring,li2022findnet}, fixation annotations \cite{lv2021simultaneously}, discriminative annotation \cite{jia2022segment}, scribble annotation \cite{he2023weakly}, depth and frequency.
    }
    \label{fig:cues}
\end{figure*}

\noindent\textbf{$\bullet$ Exploring with contrasts.}
The brain recognizes and understands different visual patterns and features by comparing and contrasting them.
By introducing multi-pattern data, the model is more likely to understand the difference between concealed and non-concealed, thereby reducing the probability of misclassifying other targets as concealed.

While all three concealment mechanisms have datasets that include non-concealed samples\cite{fan2020COD10K,wang2018stacked,dong2013casia}, artificial concealment counteraction approaches utilize these samples most. 
Introducing a large number of non-concealed samples, along with additional classification branches or special penalty strategies, enables models\cite{Wu_2019_CVPR,ma2023iml} to better distinguish between concealed and non-concealed samples.

In contrast, the non-concealed samples provided in the COD10K dataset have not been widely used in studies targeting biological concealed perception methods.
Some studies \cite{li2021uncertainty,yang2023finding} selected salient samples that were more distinct from the concealed samples as a proxy.
By introducing samples from SOD datasets, these methods jointly train the SOD and COD tasks, thus modeling the differences between the two and forcing the models to focus on the properties of concealment.
Additionally, Luo~\etal\cite{luo2023camdiff} implemented a refined scheme to generate category-consistent salient targets for a given sample by utilizing artificial intelligence-generated content technology, to exclude the influence of irrelevant factors such as target category. This contrastive learning approach is developed further by Guo~\etal\cite{guo2025enhancing}.

\subsubsection{Additional Cues}
\label{sec:Additional_Cues}
The perception of concealment involves inference and the integration of experiential knowledge. Introducing domain-specific priors as supplementary cues can provide the model with human-like guidance, improving its ability to distinguish concealed targets.
\figref{fig:cues} shows the visualization of different annotations and additional cues.

\noindent\textbf{$\bullet$ Depth.}
Research indicates that the visual system captures visual scenes that the brain's cortex utilizes to construct depth perception based on factors such as perspective cues, subtle differences in binocular images, and other depth-related cues.
%
Indeed, it is widely acknowledged that 3D knowledge of the scene can serve as valuable additional cues for visual perception tasks \cite{wu2022source,qi20173d,zhou2021rgb,howard2012perceiving}. 
On one hand, depth perception is crucial for highlighting object boundaries\cite{kelman2008review,adams2019disruptive}, and stereoscopic perception derived from depth information can reveal structural details of objects, including those concealed within the background. Leveraging cross-modal fusion between depth information and original image features can reduce ambiguity in detecting concealed targets, thereby improving method performance\cite{xiang2021exploring,wu2022source}.
On the other hand, capturing three-dimensional scene information with depth sensors based on structured light can introduce anomalies in the depth map due to factors like reflection and refraction.
For instance, the transparency of glass may cause a region in the depth map to be missing \cite{lin2022depth}, and the reflection from a mirror may result in depth discontinuities \cite{mei2021depthmirror}.
Therefore, depth cues are often employed in perception tasks related to optical property concealment, including mirror detection \cite{mei2021depthmirror,tan2021mirror3d} and transparent object detection \cite{seib2017friend,huang2018glass,xiao2021ecnet,lin2022depth,sun2023trosd}.
These methods typically build a depth prediction branch to estimate depth information from the original input image. 
Then multimodal feature fusion modules are utilized to concentrate on specific regions for analysis, thereby achieving a precise depiction of concealed targets in the scene.

\noindent\textbf{$\bullet$ Texture.}
Biological research
defines visual texture as a subset of natural images, characterized by groups of pixels sharing certain statistical regularities. By closely mimicking the texture patterns in the background environment, targets effectively blend into their surroundings. Despite the efficacy of deep neural networks in feature extraction, they often misinterpret foreground and background elements when dealing with this visual deception strategy.

Therefore, hand-crafted texture features are integrated as additional cues to enhance the model. The definition of texture features varies across methods, including capturing distinct lines in the foreground using Canny algorithm and combining them with edge contours \cite{zhu2021inferring}, calculating the covariance matrix between features in different channels to describe interactions \cite{ren2021deep}, computing differences between the original image and one processed by bilateral filtering \cite{li2022findnet}, and analyzing pixel change gradients \cite{ji2023deep}.

In a broader sense, image frequency is also considered a type of texture cue. While not directly visible, this generalization of pixel variations in two-dimensional space exhibits specific regularities. Typically, image edges and texture details are associated with high-frequency bands, while the fundamental structure and overall intensity of the image align with medium and low frequencies. 
Wavelet transforms\cite{li2018wavelet} and discrete cosine transform \cite{zhong2022detecting} can decompose signals of various frequencies in the image. 
Correlation can be calculated after enhancing signals from different frequencies, providing guidance for recognizing concealed targets.

Additionally, suppressing high-frequency information and enhancing low-frequency information has proven effective, as edges and textures in high-frequency bands are primary deceptive elements of concealment\cite{lin2023frequency}. Given that texture information can be obtained from the input image, it is common practice to create an auxiliary task to extract texture cues, fuse these with the original features, and use the combined features for concealed object perception \cite{sun2024frequency}.

\noindent\textbf{$\bullet$ Edge.}
Edges are crucial for visual perception, and even simple outlines are sufficient for object recognition.
Edge cues mark the visual boundary between objects and their backgrounds, which arises from differences in their visual features or spatial positions, and often tends to be invisible due to concealment mechanisms.
Additionally, edge cues indirectly inform the location, shape, and size of concealed targets, significantly enhancing the model's ability to perceive object boundaries and reducing the effectiveness of concealment strategies like background matching and disruptive coloration. 
In the perception of biological concealment, the high level of matching between animals and the background introduces uncertainty in edge detection.
To mitigate this, researchers use edge cues derived directly from ground truth (GT) to supervise model training. This is achieved through an additional boundary prediction branch and a refined loss function to improve the model's focus on edges\cite{zhu2022can,ji2022fast,sun2022boundary,li2022findnet,zhou2022feature,he2023feder,lee2023boundaryaware,liu2024edge,niu2024minet,guan2025promoting}.
In optical concealment, most glass and mirrors have regular edges, so edge cues often have higher confidence.
Therefore, besides their direct use as supervision \cite{luo2021espfnet, xiao2021ecnet, cao2021fakemix, tan2022mirror}, edge cues have also been employed to assist in locating and refining segmentation results \cite{lin2020progressivemirror, he2021enhanced, he2022polar, zheng2023don}.
Additionally, considering the diversity of edge cues in assisting perception, they can not only be directly used to locate concealed targets but also combined with other cues to construct composite supervision graphs, providing the model with richer guidance \cite{zhu2021inferring,jia2022segment}.

\noindent\textbf{$\bullet$ Others.}
Additional cues are often designed for specific tasks and object categories. Based on their sources and modes of operation, these cues can be categorized into two types, \ie, physical and perception cues.

Physical cues: Some objects possess unique physical properties \cite{kadambi2023incorporating} that can be used as additional cues.
Human acquisition of visual information relies on the interaction between light and matter. For transparent objects, the reduced number of effective visual features requires a deeper exploration of their interaction with light.
For example, the reflection and highlights on the glass provide positional information. Pre-extracting reflections in images can aid in locating glass \cite{lin2021rich, sudo2021Detection}. 
Differences shown by transparent objects in the light field \cite{maeno2013light, tsai2019distinguish, xu2015transcut} and polarization images \cite{mei2022rgbpglass} serve as supplementary identification information. 
%
Furthermore, due to the significant differences in the transmission of thermal energy through glass compared to other common objects, using paired RGB and thermal images makes it considerably easier to detect glass \cite{huo2023rgbt}.

Perception cues: Some tasks are closely related to perception, making additional information based on biological features or high-level semantics valuable.
Fixation data directly records the distribution of attention when people observe images, measuring the degree of object concealment from a human perception perspective and has been used in detecting concealed animals \cite{lv2021simultaneously}.
Researchers have also noted that objects often appear in specific scenes, and reasoning with the semantic information in the environment contributes to object recognition \cite{guan2022learning,lin2022exploiting}.

In conclusion, the introduction of additional cues can effectively enhance the model's perception of concealed targets. However, the integration of multimodal information and the difficulty in obtaining specific additional cues present new challenges for these solutions.

\subsection{From Distraction to Concentration}
\label{sec:From_Distraction_to_Concentration}
In most concealment categories, even after pinpointing a concealed target's location, fully isolating it from the background remains challenging.
This difficulty arises because other features in the scene often attract more attention than the target's outline. 
This issue is not only due to the inconspicuous boundaries of the target but is also exacerbated by conspicuous markings within certain concealed targets or areas, which divert attention.
This phenomenon has led researchers to explore strategies for eliminating interference and accurately attributing ambiguous regions.

\subsubsection{Elaborate Loss}
\label{sec:Elaborate_Loss}
Most image segmentation models utilize cross-entropy (CE) loss  to supervise the training process due to its simplicity and effectiveness.
The CE loss tends to apply heavier penalties for highly incorrect predictions, leading models to adopt a compromise strategy when dealing with challenging samples \cite{qin2021boundary}. 
This results in the prediction of the boundary region of an object with a value around 0.5, causing blurring.
To redirect the model's focus on edge,
Qin~\etal \cite{qin2021boundary} characterized the blurred regions as the boundary between the foreground and background, as well as the intricate structure details of the foreground objects. 
They developed a hybrid loss function combining BCE loss, SSIM loss, and IOU loss to improve the model's recognition of these areas.
Likewise, to detect subtle distinctions in blurred regions and differentiate between the background and foreground, Ren~\etal \cite{ren2021deep} introduced affinity loss to accentuate texture variances and used boundary consistency loss to reassess and refine blurred boundaries, thereby enhancing boundary quality.
This approach enhances the identification of specific regions through supervised training. However, it is tailored for particular fuzzy regions and does not adapt to changes.

\subsubsection{Iterative Refinement}
\label{sec:Iterative_Refinement}
Many concealed targets may not be initially detected. 
Instead of a single glance, repeatedly selecting target areas is another way for detection.
And each additional round reduces the likelihood of missing concealed targets. 
Jia~\etal~\cite{jia2022segment} incorporated an iterative refinement module into a multi-stage detection network, allowing samples to progressively enhance prediction accuracy. 
Similarly, Hu~\etal~\cite{hu2023high} used an iterative fusion of multi-scale features in a perception method for concealed targets, incorporating feedback after each iteration to improve high-resolution image cues. Nevertheless, repeated iteration may increase computational demands, potentially affecting the model’s lightweight design and real-time processing capabilities.

\begin{figure*}[t!]
    \centering
    \centering
	\subfloat[Element-wise multiplication]{\includegraphics[width=.22\linewidth]{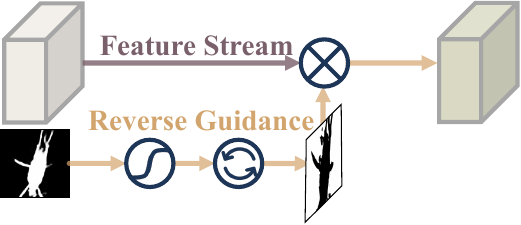}}\hspace{15pt}
	\subfloat[Addition and subtraction]{\includegraphics[width=.22\linewidth]{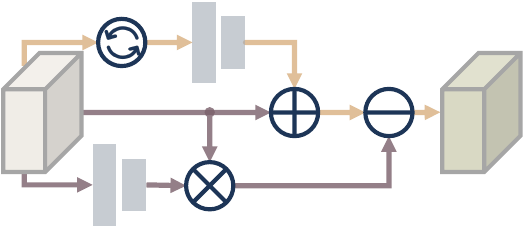}}\hspace{15pt}
	\subfloat[Channel concatenation]{\includegraphics[width=.22\linewidth]{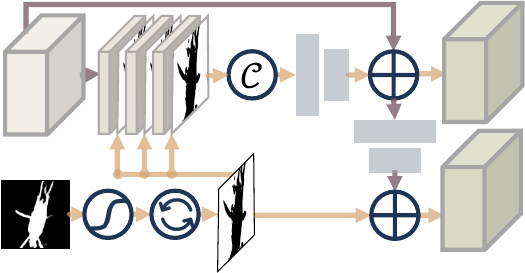}}\hspace{15pt}
        \subfloat{\includegraphics[width=.22\linewidth]{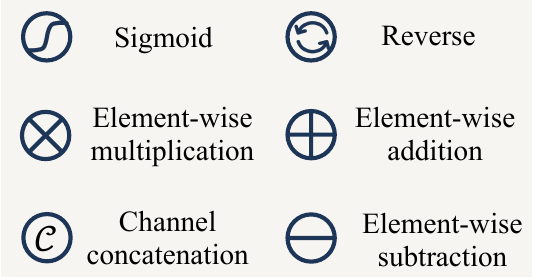}}\hspace{15pt}
    \caption{\small Illustration of three strategies of reverse guidance.
        } 
    \label{fig:reverse_guidance}
\end{figure*}

\subsubsection{Reverse Guidance}
\label{sec:Reverse_Guidance}
To distinguish ambiguous areas between background and foreground, reverse guidance has been proposed and validated as an effective strategy \cite{wei2017object,chen2018reverse}. The model can separately focus on the foreground and background by reversing intermediate prediction results. This approach prevents blending during feature refinement and accentuates boundaries. 
As shown in \figref{fig:reverse_guidance}, there are three primary methods for utilizing reverse guidance.

\noindent\textbf{$\bullet$ Element-wise multiplication.}
A straightforward approach is to reverse the coarse prediction maps and use them as attention weights \cite{chen2018reverse}. 
By applying element-wise multiplication to erase the currently predicted region, the network focuses on the background and enhances foreground details. 
Ji~\etal~\cite{ji2022fast} used this method to re-calibrate prediction map edges. 
Recognizing the importance of foreground focus, many techniques use Separate Attention strategies \cite{zhu2022can,yin2022camoformer}.
These methods employ multiple branches to concentrate on the foreground and background independently to prevent mutual contamination and then merge the two to extract more distinctive features.
%
Nevertheless, Fan~\etal\cite{fan2021concealed} argue that element-wise multiplication is constrained by the network’s discriminative capacity, making it prone to feature confusion, and simple multiplication often causes cumulative inaccuracies across iterations.

\noindent\textbf{$\bullet$ Element-wise addition and subtraction.}
Another method simulates human error correction by focusing on false positives (FP) and false negatives (FN) within the model's prediction maps and eliminating them using a dual branching structure incorporating reverse operations \cite{zheng2019distraction,mei2021camouflaged,zheng2023don}.
While element-wise multiplication is employed to focus on foreground and background features, error correction is performed through addition and subtraction operations instead. 
FP, representing incorrectly detected background regions as targets, is rectified by element-wise subtraction to suppress the blurred background. FN, representing incorrectly detected target regions as background, is corrected by element-wise addition to fill in the missing foreground. Consequently, background and foreground regions are refined separately for more discriminative features.

\noindent\textbf{$\bullet$ Channel concatenation.}
Moreover, Fan~\etal\cite{fan2021concealed} differentiated their approach by organizing features by channel and then inserting the reverse map into each group separately.
Subsequently, the acquired results are merged, representing an indirect use of reverse guidance.

Overall, reverse guidance involves directing focus toward the background with distinct attention to handling uncertain foreground and background areas separately.
\subsubsection{Uncertainty Modeling}
\label{sec:Uncertainty_Modeling}
Furthermore, another straightforward method for identifying ambiguous areas in prediction results is uncertainty modeling. Researchers define uncertainty in three ways.
Yang~\etal~\cite{yang2021uncertainty} utilized a Bayesian network to model predictions of a lightweight network by representing the output distribution of each pixel as a Gaussian distribution, predicting a stochastic component as an uncertainty map. This map guided the masking of specific regions, forcing the model to focus on these regions during prediction.
Next, Jia~\etal~\cite{jia2022segment} integrated human perception, considering edges and areas not initially attended to as ambiguous regions. They combined the dilated edge and fixation map manually to create discriminative masks, assigning weights to individual pixels within the target region to emphasize uncertain areas.
Similarly, Zheng~\etal\cite{zheng2022glassnet} and Zhou~\etal\cite{zhou2024decoupling} decoupled the mask into an uncertain boundary and an internal main body, assigning different weights to the hard and easy regions.
Finally, the disparity between ground truths (GTs) and prediction maps serves as a direct indicator of uncertainty, with the true prediction error region determined by pixel-wise subtraction. Based on this method, Zhang~\etal~\cite{zhang2022preynet} incorporated uncertainty maps into grouped features to focus on specific regions during the refinement stage, while Liu~\etal~\cite{liu2022modeling} treated confidence estimation as an additional task, acquiring uncertainty maps through additional branches.

\subsection{From Static to Dynamic}
\label{sec:From_Static_to_Dynamic}
Although various concealment mechanisms pose challenges to visual perception, they are constrained by a key limitation: movement.
%
%
The biological visual system excels at detecting objects that constantly change in appearance, posture, and location \cite{lamdouar2021segmenting}.
Even subtle concealment, like masquerade, requires animals to readjust carefully after any movement to maintain effectiveness. In other words, motion can disrupt concealment~\cite{merilaita2017howcamwork}, causing a perfectly concealed object to become instantly visible once it starts moving~\cite{bideau2016moving}.

Since the motion of concealed targets is primarily derived from the activities of concealed animals, recent efforts in video CDP have focused on biological concealment.
Bideau~\etal\cite{bideau2016moving} initially recognized the impact of motion on concealment, proceeded to detect the motion cues of concealed targets using optical flow estimation, and subsequently developed a probabilistic model to capture the three-dimensional motion details within the optical flow vector.
As a result, optical flow has become the dominant feature for depicting the motion of concealed targets and is widely used in various methods. This approach has improved from perspectives such as neuroscience~\cite{lamdouar2020betrayed}, 
data efficiency~\cite{lamdouar2021segmenting}, 
and temporal consistency~\cite{cheng2022implicit} to achieve superior segmentation outcomes.
While these studies are effective in utilizing motion cues to counteract concealment, research in this area is still in its early stages and faces challenges, including the high cost of data annotation and the difficulty of extracting motion cues in the context of concealment.
Key approaches to reduce annotation costs include developing data-efficient frameworks~\cite{meunier2022driven} and generating synthetic data~\cite{lamdouar2021segmenting}. Additionally, improving motion cue extraction methods is crucial. Chen~\etal~\cite{cheng2022implicit} have made initial strides in addressing this issue by introducing tomography and implicit motion depiction through models, underscoring the need for further exploration in accurately extracting motion indicators.

\section{CDP Benchmarks}
\subsection{Datasets and Metrics}
We conducted a review of 54 datasets used within the CDP community, covering three types of concealment. Detailed statistics, including dataset names, sources, data types, divisions, and annotation formats are presented in \tabref{tab:dataset}.

To investigate the impact of concealment on visual perception, we assess the performance of three classical dense prediction tasks (segmentation, detection, and edge estimation) on commonly used datasets.
For the concealed object segmentation task, we select four commonly used metrics: MAE ($M$)~\cite{perazzi2012saliency}, F-measure ($F_\beta$)~\cite{margolin2014evaluate}, E-measure ($E_\phi$)~\cite{fan2018enhanced,fan2021cognitive}, and Structure-measure ($S_\alpha$)~\cite{fan2017structure,cheng2021structure}.
We adhered to the COCO evaluation criteria for the concealed object detection task and selected Average Precision (AP) and its variants as the evaluation metric.
For the edge estimation task, we focus on qualitative evaluation, as ground truth annotations are not available.
%

\begin{table*}[t!]	
	\renewcommand{\arraystretch}{1.2}
	\setlength{\tabcolsep}{2pt}
        \vspace{-15pt}
        \captionsetup{format=plain,labelformat=empty}
        \caption{
            \begin{tcolorbox}[colback=captionbg, colframe=white, sharp corners=south, boxrule=0mm, width=.99\linewidth]
                \textcolor{captiontext}{
                \small \textbf{TABLE 1:}\textbf{~Essential characteristics for CDP datasets.} 
                \textbf{Available (Avail.):} whether open access.
                \textbf{Modal (Mod.):} data type of dataset.
                \textbf{Subclass (Sub.Cls.):} main categories of objects in the dataset, including transparent object (Trans.), mirror (Mirr.), shadow (Shad.), and so on.
                \textbf{Task:} tasks for which the dataset was used, which can be segmentation (Seg.), Counting (Cnt.), Tracking (Tra.), Locating (Loc.), and Matting (Mat.).
                \textbf{Synthetic Data (SYN):} Does it contain synthetic data.
                \textbf{Real Data (Real):} Does it contain real data.
                \textbf{Train/Test:} number of samples for training/testing.
                \textbf{N.C.:} whether including non-concealed samples.
                \textbf{Box.:} whether providing bounding box annotations for detection.
                \textbf{Obj.:} whether providing object-level segmentation masks.
                \textbf{Ins.:} whether including instance-level segmentation masks.
                \textbf{Adds.:} the additional annotations.
                }
            \end{tcolorbox}
        }
        \vspace{-20pt}
        \label{tab:dataset}
	\begin{center}	
	    \footnotesize
		\begin{tabular}{cclllccccccccccccc}
                \specialrule{1.2pt}{0pt}{0pt}  
                \rowcolor{headerbg} 
			  \textbf{Class} & \# & \textbf{Datasets} & \textbf{Year} & \textbf{Pub.} & \textbf{Avail.} & \textbf{Mod.} & \textbf{Sub.Cls} & \textbf{Task} & \textbf{SYN} & \textbf{Real} & \textbf{Train} & \textbf{Test} & \textbf{N.C.} & \textbf{Box.} & \textbf{Obj.} & \textbf{Ins.} & \textbf{Adds.} \\\hline
                 \rowcolor{tablebg}  & 1 & CHAMELEON\cite{datasetCHAMELEON} & 2017 & - & \href{https://www.polsl.pl/rau6/chameleon-database-animal-camouflage-analysis/}{Link} & Image & Animal & Seg. &    & \checkmark & 0 & 76 &    &    & \checkmark &    & - \\
                \rowcolor{tablebg} & 2 & CPD1K\cite{datasetCPD} & 2018 & SPL & \href{https://github.com/xfflyer/Camouflaged-Data}{Link} & Image & Human & Seg. &    & \checkmark & 600 & 400 &    &    & \checkmark &    & - \\
                \rowcolor{tablebg} & 3 & CAMO\cite{datasetCAMO} & 2019 & CVIU & \href{https://sites.google.com/view/ltnghia/research/camo}{Link} & Image & Animal & Seg. &    & \checkmark & 1000 & 250 &    &    & \checkmark &    & - \\
                \rowcolor{tablebg} & 4 & COD10K\cite{fan2020COD10K} & 2020 & CVPR & \href{https://dengpingfan.github.io/pages/COD.html}{Link} & Image & Animal & Seg. &    & \checkmark & 6000 & 4000 & \checkmark & \checkmark & \checkmark & \checkmark & - \\
                \rowcolor{tablebg} & 5 & NC4K\cite{lv2021simultaneously} & 2021 & CVPR & \href{https://github.com/JingZhang617/COD-Rank-Localize-and-Segment}{Link} & Image & Animal & Seg. &    & \checkmark & 0 & 4121 &    & \checkmark & \checkmark & \checkmark & - \\
                \rowcolor{tablebg} & 6 & CAMO++\cite{le2021camouflaged} & 2021 & TIP & - & Image & Animal & Seg. &    & \checkmark & 3500 & 2000 & \checkmark & \checkmark & \checkmark & \checkmark & - \\
                \rowcolor{tablebg} & 7 & CAM-LDR \cite{lv2023towards} & 2023 & TCSVT & \href{https://github.com/JingZhang617/COD-Rank-Localize-and-Segment}{Link} & Image & Animal & Seg. &    & \checkmark & 4040 & 2026 &    &    & \checkmark & \checkmark & Ranking \\
                \rowcolor{tablebg} & 8 & S-COD \cite{he2023weakly} & 2023 & AAAI & \href{https://github.com/dddraxxx/Weakly-Supervised-Camouflaged-Object-Detection-with-Scribble-Annotations}{Link} & Image & Animal & Seg. &    & \checkmark & 4040 & 0 &    &    &    &    & Scribble \\
                \rowcolor{tablebg} & 9 & IOCfish5K \cite{sun2023iocformer} & 2023 & CVPR & \href{https://github.com/GuoleiSun/Indiscernible-Object-Counting}{Link} & Image & Fish & Cnt. &    & \checkmark & 3637 & 2000 &    &    &    &    & Count \\
                \rowcolor{tablebg} & 10 & CoCOD8K \cite{zhang2023collaborative} & 2023 & TNNLS & \href{https://github.com/zc199823/BBNet--CoCOD}{Link} & Image & Animal & Seg. &    & \checkmark & 5933 & 2595 &    &    & \checkmark &    & Collaborative \\
                \rowcolor{tablebg} & 11 & ACOD2K \cite{song2023camouflaged} & 2023 & ICME & \href{https://github.com/syxvision/FDNet}{Link} & Image & Human & Seg. &    & \checkmark & 1600 & 400 & \checkmark &    & \checkmark &    & - \\
                \rowcolor{tablebg} & 12 & R2C7K \cite{zhang2023referring} & 2023 & arXiv & \href{https://github.com/zhangxuying1004/RefCOD}{Link} & Image & Animal & Seg. &    & \checkmark &2979  &1987  & \checkmark &    & \checkmark &    & Reference \\
                \rowcolor{tablebg} & 13 & ACOD12K \cite{wang2024depth} & 2024 & CVPR & \href{https://github.com/Kki2Eve/RISNet}{Link} & Image & Crop & Seg. &    & \checkmark & 4600 & 1492 &    &    & \checkmark &    & Depth \\
                \rowcolor{tablebg} & 14 & P-COD \cite{chen2024just} & 2024 & ECCV & \href{https://github.com/2231122/PCOD}{Link} & Image & Animal & Seg. &    & \checkmark & 4040 & 0 &    &    &    &    & Point \\
                \rowcolor{tablebg} & 15 & COD-TAX \cite{zhang2024unlocking} & 2024 & ECCV & \href{https://github.com/lyu-yx/ACUMEN}{Link} & Image & Animal & Seg. &    & \checkmark & 4040 & 0 &    &    &    &    & Text \\
                \rowcolor{tablebg} & 16 & OVCamo \cite{pang2024open} & 2024 & ECCV & \href{https://github.com/lartpang/OVCamo}{Link} & Image & Mixed & Seg. &    & \checkmark & 7713 & 3770 &    &    &  \checkmark  &    & Category \\
                \rowcolor{tablebg} & 17 & PlantCamo \cite{plantcamo} & 2025 & AIR & \href{https://github.com/yjybuaa/PlantCamo}{Link} & Image & Plant & Seg. &    & \checkmark & 1000 & 250 &    &    & \checkmark &    &  \\
                \rowcolor{tablebg} & 18 & CAD\cite{bideauECCV16} & 2016 & ECCV & \href{http://vis-www.cs.umass.edu/motionSegmentation/}{Link} & Video & Animal & Tra &    & \checkmark & 0 & 836 &    &    & \checkmark &    & - \\
                \rowcolor{tablebg} & 19 & MoCA\cite{lamdouar2020MoCA} & 2020 & ACCV & \href{https://www.robots.ox.ac.uk/~vgg/data/MoCA/}{Link} & Video & Animal & Tra &    & \checkmark & 0 & 37250 &    & \checkmark &    &    & - \\
                \rowcolor{tablebg} \multirow{-16}{*}{\rotatebox{90}{\makecell[c]{Biological \\ Concealment}}} & 20 & MoCA-Mask\cite{cheng2022MoCAMask} & 2022 & CVPR & \href{https://xueliancheng.github.io/SLT-Net-project/}{Link} & Video & Animal & Tra &    & \checkmark & 19313 & 3626 &    &    & \checkmark &    & - \\
                 \specialrule{0pt}{1pt}{1pt}
                \rowcolor{tablebg} & 21 & Transcut \cite{xu2015transcut} & 2015 & ICCV & - & Image & Trans. & Seg. &    & \checkmark & 0 & 49 &    &    & \checkmark &    & Light Filed \\ 
                \rowcolor{tablebg} & 22 & Seib V~\etal\cite{seib2017friend} & 2017 & ICRMV & \href{http://agas.uni-koblenz.de/data/datasets/transparent_objects/transparent_objects.tar.gz}{Link} & Image & Trans. & Loc. &    & \checkmark & 3618 & 792 & \checkmark & \checkmark &    &    & Depth \\
                \rowcolor{tablebg} & 23 & TomNet\cite{chen2018tomnet} & 2018 & CVPR & \href{https://github.com/guanyingc/TOM-Net}{Link} & Image & Trans. & Mat. & \checkmark & \checkmark & 178k & 876 &    &    & \checkmark &    & - \\
                \rowcolor{tablebg} & 24 & GDD\cite{meidonhitme} & 2020 & CVPR & \href{https://mhaiyang.github.io/CVPR2020_GDNet/index.html}{Link} & Image & Trans. & Seg. &    & \checkmark & 2980 & 936 &    &    & \checkmark &    & - \\
                \rowcolor{tablebg} & 25 & Kalra$\sim$\etal\cite{kalra2020deep} & 2020 & CVPR & - & Image & Trans. & Seg. &    & \checkmark & 1000 & 600 &    &    & \checkmark & \checkmark & Polarization \\
                \rowcolor{tablebg} & 26 & Trans10K\cite{xie2020trans10k}\cite{xie2021segmenting} & 2020 & ECCV & \href{https://github.com/xieenze/Segment_Transparent_Objects}{Link} & Image & Trans. & Seg. &    & \checkmark & 5000 & 4428 &    &    & \checkmark &    & - \\
                \rowcolor{tablebg} & 27 & ClearGrasp\cite{sajjan2020cleargrasp} & 2020 & ICRA & \href{https://github.com/xieenze/Segment_Transparent_Objects}{Link} & Image & Trans. & Seg. & \checkmark & \checkmark & 50k & 286 &    &    & \checkmark &    & Depth \\
                \rowcolor{tablebg} & 28 & GSD\cite{lin2021rich} & 2021 & CVPR & \href{https://jiaying.link/cvpr2021-gsd/}{Link} & Image & Trans. & Seg. &    & \checkmark & 3202 & 810 &    &    & \checkmark &    & - \\
                \rowcolor{tablebg} & 29 & Window\cite{zhu2021global} & 2021 & ICIG & - & Image & Trans. & Seg. &    & \checkmark & 0 & 4419 &    &    & \checkmark &    & - \\
                \rowcolor{tablebg} & 30 & ATOM\cite{fan2021transparent} & 2021 & ICCV & - & Image & Trans. & Tra. &    & \checkmark & 0 & 86k &    & \checkmark &    &    & - \\
                \rowcolor{tablebg} & 31 & RGBD-GSD\cite{lin2022rgbdgsd} & 2022 & arXiv & - & Image & Trans. & Seg. &    & \checkmark & 2400 & 609 &    &    & \checkmark &    & Depth \\
                \rowcolor{tablebg} & 32 & RGBP-Glass\cite{mei2022rgbpglass} & 2022 & CVPR & \href{https://mhaiyang.github.io/CVPR2022_PGSNet/index.html}{Link} & Image & Trans. & Seg. &    & \checkmark & 3207 & 1304 &    & \checkmark & \checkmark &    & Polarization \\
                \rowcolor{tablebg} & 33 & GSD-S\cite{lin2022RGBS} & 2022 & NeurIPS & \href{https://jiaying.link/neurips2022-gsds/}{Link} & Image & Trans. & Seg. &    & \checkmark & 3911 & 608 &    &    & \checkmark &    & Semantic \\
                \rowcolor{tablebg} & 34 & HSO\cite{yu2022progressive} & 2022 & TIP & \href{https://mhaiyang.github.io/TIP2022-PGSNet/index.html}{Link} & Image & Trans. & Seg. &    & \checkmark & 3070 & 1782 &    &    & \checkmark &    & - \\
                \rowcolor{tablebg} & 35 & Trans2K\cite{lukezic2022trans2k} & 2022 & arXiv & \href{https://github.com/trojerz/Trans2k}{Link} & Image & Trans. & Tra. & \checkmark &    & 104343 & 0 &    & \checkmark & \checkmark &    & - \\
                \rowcolor{tablebg} & 36 & ClearPose\cite{chen2022clearpose} & 2022 & ECCV & \href{https://github.com/opipari/ClearPose}{Link} & Image & Trans. & Seg. &    & \checkmark &354481  & 0 &    &    & \checkmark &    & Depth \\
                \rowcolor{tablebg} & 37 & Huo Dong\etal\cite{huo2023rgbt} & 2023 & TIP & \href{https://github.com/Dong-Huo/RGB-T-Glass-Segmentation}{Link} & Image & Trans. & Seg. &    & \checkmark & 4427 & 1124 & \checkmark &    & \checkmark &    & Thermal \\
                \rowcolor{tablebg} & 38 & TROSD\cite{sun2023trosd} & 2023 & TCSVT & \href{http://www.tsinghua-ieit.com/trosd}{Link} & Image & Trans. & Seg. &    & \checkmark & 6748 & 2432 &    &    & \checkmark &    & - \\
                \rowcolor{tablebg} & 39 & MSD\cite{yang2019my} & 2019 & ICCV & \href{https://mhaiyang.github.io/ICCV2019_MirrorNet/index.html}{Link} & Image & Mirr. & Seg. &    & \checkmark & 3063 & 955 &    &    & \checkmark &    & - \\
                \rowcolor{tablebg} & 40 & PMD\cite{lin2020progressivemirror} & 2020 & CVPR & \href{https://jiaying.link/cvpr2020-pgd/}{Link} & Image & Mirr. & Seg. &    & \checkmark & 5096 & 1365 &    &    & \checkmark &    & - \\
                \rowcolor{tablebg} & 41 & RGBD-Mirror\cite{mei2021depthmirror} & 2021 & CVPR & \href{https://mhaiyang.github.io/CVPR2021_PDNet/index}{Link} & Image & Mirr. & Seg. &    & \checkmark & 2000 & 1049 &    &    & \checkmark &    & Depth \\
                \rowcolor{tablebg} & 42 & NMD\cite{he2022NMD} & 2022 & IVC & \href{}{Link} & Image & Mirr. & Seg. &    & \checkmark & 4285 & 1440 & \checkmark &    &    &    & -  \\
                \rowcolor{tablebg} & 43 & UCF \cite{zhu2010learning} & 2010 & CVPR & \href{https://github.com/eraserNut/MTMT?tab=readme-ov-file}{Link} & Image & Shad. & Seg. &    & \checkmark & 125 & 120 & \checkmark &    & \checkmark &    & - \\
                \rowcolor{tablebg} & 44 & UIUC \cite{guo2011single} & 2011 & CVPR & - & Image & Shad. & Seg. &    & \checkmark & 32 & 76 & \checkmark &    & \checkmark &    & - \\
                \rowcolor{tablebg} & 45 & SBU \cite{vicente2016large} & 2016 & ECCV & \href{http://www3.cs.stonybrook.edu/~cvl/dataset.html.}{Link} & Image & Shad. & Seg. &    & \checkmark & 4089 & 638 &    &    & \checkmark &    & - \\
                \rowcolor{tablebg} & 46 & ISTD \cite{wang2018stacked} & 2018 & CVPR & \href{https://github.com/DeepInsight-PCALab/ST-CGAN}{Link} & Image & Shad. & Seg. &    &    & 1330 & 540 & \checkmark &    & \checkmark &    & - \\
                \rowcolor{tablebg} & 47 & CUHK-Shadow \cite{hu2021revisiting} & 2021 & TIP & \href{https://github.com/xw-hu/CUHK-Shadow}{Link} & Image & Shad. & Seg. &    & \checkmark & 7350 & 2100 &    &    & \checkmark &    & - \\
               \rowcolor{tablebg} \multirow{-28}{*}{\rotatebox{90}{\makecell[c]{Optical Property \\ Concealment}}} & 48 & SOBA \cite{wang2022instance} & 2022 & TPAMI & \href{https://github.com/stevewongv/SSIS}{Link} & Image & Shad. & Seg. &    &    & 840 & 260 &    &    & \checkmark & \checkmark & - \\
                 \specialrule{0pt}{1pt}{1pt}
               \rowcolor{tablebg} & 49 & Columbia \cite{hsu2006columbia} & 2006 & - & \href{https://www.ee.columbia.edu/ln/dvmm/downloads/AuthSplicedDataSet/AuthSplicedDataSet.htm}{Link} & Image & Tamper & Seg. & \checkmark &    & 0 & 263 &    &    & \checkmark &    & -\\
               \rowcolor{tablebg} & 50 & CASIA-v1 \cite{dong2013casia} & 2013 & ChinaSIP & \href{https://link.zhihu.com/?target=https%3A//github.com/namtpham/casia1groundtruth}{Link} & Image & Tamper & Seg. & \checkmark &    & 0 & 1721 & \checkmark &    & \checkmark &    & - \\
                \rowcolor{tablebg} & 51 & CASIA-v2 \cite{dong2013casia} & 2013 & ChinaSIP & \href{https://link.zhihu.com/?target=https%3A//github.com/namtpham/casia2groundtruth}{Link} & Image & Tamper & Seg. & \checkmark &    & 12323 & 0 & \checkmark &    & \checkmark &    & - \\
                \rowcolor{tablebg} & 52 & COVER \cite{wen2016coverage} & 2016 & ICIP & \href{https://1drv.ms/f/s!AggVhXcCj1FLhUUyUrqSpV_yI_GH}{Link} & Image & Tamper & Seg. & \checkmark &    & 0 & 200 & \checkmark &    & \checkmark &    & -\\     
                \rowcolor{tablebg} & 53 & NIST16 \cite{guan2019mfc} & 2019 & WACVW & \href{https://mfc.nist.gov/}{Link} & Image & Tamper & Seg. & \checkmark &    & 404 & 160 &    &    & \checkmark &    & -\\       
                \rowcolor{tablebg} \multirow{-6}{*}{\rotatebox{90}{\makecell[c]{Artificial \\ Concealment}}} & 54 & IMD2020 \cite{novozamsky2020imd2020} & 2020 & WACVW & \href{http://staff.utia.cas.cz/novozada/db}{Link} & Image & Tamper & Seg. & \checkmark &    & 0 & 2424 &    &    & \checkmark &    & -\\
                \specialrule{1.2pt}{0pt}{0pt}
		\end{tabular}
	\end{center}
    \vspace{-15pt}
\end{table*}

\subsection{Effectiveness of Concealment Counteraction}

\begin{table*}[ht]
    \scriptsize
    \renewcommand{\arraystretch}{1.2}
    \setlength{\tabcolsep}{1.6pt}
    \centering
    \captionsetup{format=plain,labelformat=empty}
    \caption{
        \begin{tcolorbox}[colback=captionbg, colframe=white, sharp corners=south, boxrule=0mm, width=.99\linewidth]
            \small \textbf{TABLE 2:}\textbf{~Effectiveness comparison of concealment counteraction in COS task.} 
            CH: CHAMELEON \cite{datasetCHAMELEON}, CA: CAMO \cite{datasetCAMO}, CO: COD10K \cite{fan2020COD10K}, NC: NC4K \cite{lv2021simultaneously},TR: Trans10K \cite{xie2020trans10k}, GS: GSD \cite{lin2021rich}, PM: PMD \cite{lin2020progressivemirror}, MS: MSD \cite{yang2019my}, SB: SBU \cite{vicente2016large}, IS: ISTD \cite{wang2018stacked}, CS: CASIA \cite{dong2013casia}, CV: COVER \cite{wen2016coverage}.
            The results are represented by weight F-measure ($wF_\beta$), average E-measure ($aE_\phi$), Structure-measure ($S_\alpha$), and MAE ($M$).                        
        \end{tcolorbox}
    }
    \vspace{-5pt}
    \begin{tabular}{cc}
        \belowrulesep=0pt
        \aboverulesep=0pt
        \begin{minipage}{0.5\linewidth}
            \centering
            \subfloat[$wF_\beta\uparrow$]{
                \begin{tabular}{lcccccccccccc}
                    \specialrule{1.2pt}{0pt}{0pt}
                    \rowcolor{headerbg} \textbf{Methods} & \textbf{CH} & \textbf{CA} & \textbf{CO} & \textbf{NC} & \textbf{TR} & \textbf{GS} & \textbf{PM} & \textbf{MS} & \textbf{SB} & \textbf{IS} & \textbf{CS} & \textbf{CV} \\\hline
                    SINet \cite{fan2020COD10K} & .649 & .564 & .482 & .611 & .142 & .814 & .654 & .771 & .788 & .820 & .408 & .119 \\
                    SINet-V2 \cite{fan2021concealed} & .806 & .742 & .669 & .762 & .165 & .854 & .742 & .796 & .870 & .920 & .550 & .185 \\
                    ZoomNet \cite{pang2022zoom} & .800 & .701 & .695 & .765 & .170 & .833 & .692 & .746 & .870 & .844 & .447 & .183 \\
                    HitNet \cite{hu2023high} & .703 & .677 & .666 & .757 & .162 & .843 & .504 & .760 & .851 & .933 & .472 & .121 \\
                    FEDER \cite{he2023camouflaged} & .833 & .739 & .712 & .788 & .174 & .852 & .726 & .837 & .879 & .936 & .496 & .135 \\
                    FSPNet \cite{huang2023feature} & .840 & .791 & .727 & .812 & .174 & .875 & .730 & .881 & .881 & .865 & .666 & .251 \\
                    \specialrule{0pt}{1pt}{1pt}
                    RFENet \cite{fan2023rfenet} & .848 & .755 & .710 & .795 & .174 & .870 & .711 & .837 & .871 & .903 & .462 & .135 \\
                    HetNet \cite{he2023efficient} & .828 & .739 & .701 & .792 & .173 & .856 & .734 & .841 & .872 & .903 & .477 & .099 \\
                    SATNet \cite{huang2023symmetry} & .786 & .788 & .708 & .797 & .927 & .864 & .678 & .868 & .878 & .949 & .421 & .073 \\
                    BDRAR \cite{zhu2018bidirectional} & .646 & .627 & .491 & .637 & .882 & .804 & .607 & .709 & .807 & .839 & .427 & .169 \\
                    MTMTNet \cite{chen2020multi} & .494 & .494 & .389 & .541 & .896 & .800 & .531 & .708 & .830 & .897 & .343 & .145 \\
                    FDRNet \cite{zhu2021mitigating} & .508 & .504 & .354 & .497 & .899 & .821 & .453 & .605 & .601 & .887 & .282 & .150 \\
                    SDDNet \cite{cong2023sddnet} & .548 & .538 & .422 & .564 & .897 & .814 & .559 & .590 & .700 & .870 & .382 & .179 \\
                    \specialrule{0pt}{1pt}{1pt}
                    ManTraNet~\cite{Wu_2019_CVPR} & .192 & .233 & .131 & .206 & .085 & .437 & .116 & .237 & .628 & .636 & .096 & .106 \\
                    IML-ViT~\cite{ma2023iml} & .662 & .681 & .710 & .765 & .108 & .865 & .644 & .807 & .846 & .895 & .595 & .209 \\
                    \specialrule{0pt}{1pt}{1pt}
                    SAM~\cite{kirillov2023segment} & .652 & .594 & .700 & .696 & .703 & .609 & .747 & .645 & .544 & .647 & .702 & .832 \\
                    SAM-BBox~\cite{kirillov2023segment} & .828 & .864 & .859 & .876 & .901 & .904 & .869 & .766 & .733 & .780 & .872 & .870 \\
                    SAM2~\cite{ravi2024sam2} & .597 & .612 & .617 & .708 & .430 & .219 & .115 & .240 & .265 & .307 & .325 & .348 \\
                    SAM2-BBox~\cite{ravi2024sam2} & .888 & .891 & .903 & .919 & .924 & .902 & .855 & .778 & .751 & .780 & .889 & .866 \\
                    UniverSeg~\cite{butoi2023universeg} & .139 & .186 & .098 & .162 & .042 & .246 & .091 & .184 & .114 & .100 & .099 & .136 \\
                    SegGPT~\cite{wang2023seggpt} & .380 & .318 & .373 & .416 & .043 & .293 & .271 & .171 & .077 & .140 & .046 & .040 \\
                    \bottomrule[1pt]
                \end{tabular}
            }
        \end{minipage} &
        \belowrulesep=0pt
        \aboverulesep=0pt
        \begin{minipage}{0.5\linewidth}
            \centering
            \subfloat[$aE_\phi\uparrow$]{
                \begin{tabular}{lcccccccccccc}
                    \specialrule{1.2pt}{0pt}{0pt}
                    \rowcolor{headerbg} \textbf{Methods} & \textbf{CH} & \textbf{CA} & \textbf{CO} & \textbf{NC} & \textbf{TR} & \textbf{GS} & \textbf{PM} & \textbf{MS} & \textbf{SB} & \textbf{IS} & \textbf{CS} & \textbf{CV} \\\hline
                    SINet \cite{fan2020COD10K} & .839 & .745 & .764 & .805 & .780 & .877 & .838 & .870 & .903 & .924 & .649 & .445 \\
                    SINet-V2 \cite{fan2021concealed} & .940 & .875 & .881 & .899 & .806 & .905 & .879 & .885 & .948 & .970 & .788 & .567 \\
                    ZoomNet \cite{pang2022zoom} & .919 & .830 & .864 & .881 & .743 & .884 & .844 & .835 & .932 & .907 & .675 & .513 \\
                    HitNet \cite{hu2023high} & .844 & .800 & .848 & .872 & .832 & .893 & .674 & .837 & .927 & .973 & .571 & .408 \\
                    FEDER \cite{he2023camouflaged} & .940 & .872 & .899 & .908 & .741 & .902 & .873 & .903 & .951 & .975 & .756 & .456 \\
                    FSPNet \cite{huang2023feature} & .934 & .899 & .892 & .915 & .741 & .916 & .863 & .932 & .946 & .931 & .813 & .546 \\
                    \specialrule{0pt}{1pt}{1pt}
                    RFENet \cite{fan2023rfenet} & .958 & .879 & .895 & .910 & .740 & .918 & .845 & .904 & .949 & .955 & .666 & .456 \\
                    HetNet \cite{he2023efficient} & .943 & .868 & .894 & .911 & .739 & .906 & .868 & .909 & .947 & .955 & .688 & .425 \\
                    SATNet \cite{huang2023symmetry} & .912 & .899 & .875 & .906 & .960 & .908 & .803 & .926 & .948 & .982 & .619 & .324 \\
                    BDRAR \cite{zhu2018bidirectional} & .856 & .812 & .781 & .838 & .938 & .875 & .810 & .829 & .923 & .925 & .700 & .547 \\
                    MTMTNet \cite{chen2020multi} & .727 & .683 & .672 & .755 & .944 & .867 & .754 & .825 & .925 & .955 & .622 & .517 \\
                    FDRNet \cite{zhu2021mitigating} & .764 & .727 & .678 & .742 & .948 & .890 & .762 & .759 & .760 & .956 & .647 & .540 \\
                    SDDNet \cite{cong2023sddnet} & .787 & .747 & .732 & .789 & .946 & .884 & .791 & .753 & .843 & .951 & .671 & .566 \\
                    \specialrule{0pt}{1pt}{1pt}
                    ManTraNet~\cite{Wu_2019_CVPR} & .416 & .406 & .393 & .396 & .598 & .515 & .454 & .476 & .794 & .804 & .413 & .414 \\
                    IML-ViT~\cite{ma2023iml} & .863 & .821 & .880 & .879 & .690 & .910 & .811 & .884 & .931 & .954 & .790 & .521 \\
                    \specialrule{0pt}{1pt}{1pt}
                    SAM~\cite{kirillov2023segment} & .737 & .681 & .798 & .774 & .759 & .656 & .835 & .721 & .667 & .753 & .785 & .911 \\
                    SAM-BBox~\cite{kirillov2023segment} & .911 & .924 & .947 & .940 & .919 & .895 & .925 & .832 & .861 & .890 & .930 & .949 \\
                    SAM2~\cite{ravi2024sam2} & .798 & .783 & .821 & .852 & .628 & .459 & .540 & .518 & .560 & .529 & .621 & .638 \\
                    SAM2-BBox~\cite{ravi2024sam2} & .949 & .939 & .972 & .967 & .940 & .895 & .916 & .837 & .871 & .903 & .909 & .958 \\
                    UniverSeg~\cite{butoi2023universeg} & .385 & .436 & .377 & .428 & .624 & .452 & .416 & .496 & .421 & .388 & .417 & .495 \\
                    SegGPT~\cite{wang2023seggpt} & .563 & .502 & .584 & .590 & .805 & .464 & .532 & .384 & .327 & .365 & .317 & .320 \\
                    \bottomrule[1pt]
                \end{tabular}
            }
        \end{minipage} \\
        \vspace{-15pt} \\
        \begin{minipage}{0.5\linewidth}
            \belowrulesep=0pt
            \aboverulesep=0pt
            \centering
            \subfloat[$S_\alpha\uparrow$]{
                \begin{tabular}{lcccccccccccc}
                    \specialrule{1.2pt}{0pt}{0pt}
                    \rowcolor{headerbg} \textbf{Methods} & \textbf{CH} & \textbf{CA} & \textbf{CO} & \textbf{NC} & \textbf{TR} & \textbf{GS} & \textbf{PM} & \textbf{MS} & \textbf{SB} & \textbf{IS} & \textbf{CS} & \textbf{CV} \\\hline
                    SINet \cite{fan2020COD10K} & .839 & .748 & .748 & .792 & .801 & .827 & .789 & .831 & .853 & .896 & .654 & .482 \\
                    SINet-V2 \cite{fan2021concealed} & .887 & .820 & .811 & .844 & .813 & .855 & .835 & .853 & .877 & .943 & .731 & .499 \\
                    ZoomNet \cite{pang2022zoom} & .872 & .789 & .819 & .841 & .740 & .838 & .812 & .819 & .869 & .887 & .667 & .510 \\
                    HitNet \cite{hu2023high} & .801 & .767 & .793 & .828 & .833 & .840 & .703 & .818 & .852 & .941 & .685 & .481 \\
                    FEDER \cite{he2023camouflaged} & .888 & .802 & .822 & .848 & .738 & .851 & .821 & .870 & .874 & .950 & .681 & .481 \\
                    FSPNet \cite{huang2023feature} & .904 & .854 & .847 & .877 & .740 & .869 & .833 & .906 & .884 & .908 & .801 & .554 \\
                    \specialrule{0pt}{1pt}{1pt}
                    RFENet \cite{fan2023rfenet} & .890 & .812 & .816 & .848 & .736 & .868 & .809 & .869 & .870 & .925 & .676 & .481 \\
                    HetNet \cite{he2023efficient} & .890 & .808 & .820 & .855 & .736 & .854 & .827 & .871 & .873 & .927 & .686 & .471 \\
                    SATNet \cite{huang2023symmetry} & .844 & .829 & .805 & .843 & .916 & .855 & .786 & .884 & .864 & .952 & .642 & .464 \\
                    BDRAR \cite{zhu2018bidirectional} & .826 & .784 & .746 & .807 & .908 & .826 & .784 & .798 & .853 & .889 & .687 & .486 \\
                    MTMTNet \cite{chen2020multi} & .705 & .670 & .654 & .728 & .924 & .837 & .731 & .804 & .863 & .932 & .616 & .480 \\
                    FDRNet \cite{zhu2021mitigating} & .732 & .698 & .643 & .713 & .923 & .839 & .696 & .710 & .710 & .924 & .575 & .473 \\
                    SDDNet \cite{cong2023sddnet} & .765 & .732 & .703 & .768 & .922 & .845 & .753 & .718 & .792 & .922 & .635 & .491 \\
                    \specialrule{0pt}{1pt}{1pt}
                    ManTraNet~\cite{Wu_2019_CVPR} & .483 & .469 & .439 & .460 & .610 & .550 & .449 & .453 & .779 & .777 & .454 & .447 \\
                    IML-ViT~\cite{ma2023iml} & .816 & .814 & .859 & .880 & .663 & .878 & .809 & .868 & .885 & .929 & .786 & .528 \\
                    \specialrule{0pt}{1pt}{1pt}
                    SAM~\cite{kirillov2023segment} & .720 & .678 & .782 & .763 & .733 & .620 & .797 & .688 & .581 & .683 & .771 & .871 \\
                    SAM-BBox~\cite{kirillov2023segment} & .854 & .863 & .888 & .887 & .880 & .852 & .872 & .766 & .752 & .807 & .887 & .903 \\
                    SAM2~\cite{ravi2024sam2} & .717 & .705 & .752 & .788 & .547 & .383 & .457 & .444 & .430 & .448 & .547 & .547 \\
                    SAM2-BBox~\cite{ravi2024sam2} & .900 & .890 & .920 & .921 & .904 & .857 & .864 & .781 & .770 & .807 & .903 & .901 \\
                    UniverSeg~\cite{butoi2023universeg} & .341 & .371 & .361 & .379 & .609 & .374 & .375 & .374 & .300 & .280 & .378 & .412 \\
                    SegGPT~\cite{wang2023seggpt} & .614 & .562 & .627 & .629 & .820 & .474 & .566 & .458 & .426 & .471 & .461 & .446 \\
                    \bottomrule[1pt]
                \end{tabular}
            }
        \end{minipage} &
        \begin{minipage}{0.5\linewidth}
            \belowrulesep=0pt
            \aboverulesep=0pt
            \centering
            \subfloat[$M\downarrow$]{
                \begin{tabular}{lcccccccccccc}
                    \specialrule{1.2pt}{0pt}{0pt}
                    \rowcolor{headerbg} \textbf{Methods} & \textbf{CH} & \textbf{CA} & \textbf{CO} & \textbf{NC} & \textbf{TR} & \textbf{GS} & \textbf{PM} & \textbf{MS} & \textbf{SB} & \textbf{IS} & \textbf{CS} & \textbf{CV} \\\hline
                    SINet \cite{fan2020COD10K} & .059 & .112 & .063 & .079 & .178 & .076 & .039 & .075 & .044 & .040 & .080 & .140 \\
                    SINet-V2 \cite{fan2021concealed} & .032 & .070 & .038 & .048 & .168 & .057 & .030 & .063 & .029 & .016 & .073 & .127 \\
                    ZoomNet \cite{pang2022zoom} & .029 & .079 & .032 & .047 & .236 & .071 & .032 & .074 & .030 & .055 & .088 & .125 \\
                    HitNet \cite{hu2023high} & .045 & .078 & .034 & .048 & .141 & .062 & .041 & .064 & .032 & .014 & .057 & .113 \\
                    FEDER \cite{he2023camouflaged} & .029 & .072 & .032 & .043 & .239 & .062 & .032 & .054 & .027 & .013 & .083 & .122 \\
                    FSPNet \cite{huang2023feature} & .024 & .053 & .027 & .036 & .238 & .050 & .026 & .034 & .026 & .034 & .039 & .106 \\
                    \specialrule{0pt}{1pt}{1pt}
                    RFENet \cite{fan2023rfenet} & .025 & .066 & .031 & .042 & .241 & .050 & .034 & .046 & .028 & .022 & .067 & .122 \\
                    HetNet \cite{he2023efficient} & .028 & .071 & .033 & .041 & .241 & .056 & .029 & .049 & .029 & .021 & .071 & .119 \\
                    SATNet \cite{huang2023symmetry} & .032 & .059 & .028 & .038 & .031 & .053 & .031 & .040 & .027 & .010 & .084 & .121 \\
                    BDRAR \cite{zhu2018bidirectional} & .064 & .101 & .069 & .074 & .043 & .074 & .047 & .085 & .041 & .037 & .114 & .169 \\
                    MTMTNet \cite{chen2020multi} & .129 & .185 & .122 & .119 & .038 & .077 & .073 & .102 & .038 & .021 & .183 & .203 \\
                    FDRNet \cite{zhu2021mitigating} & .112 & .154 & .121 & .125 & .037 & .068 & .083 & .139 & .140 & .021 & .222 & .233 \\
                    SDDNet \cite{cong2023sddnet} & .102 & .139 & .092 & .099 & .038 & .069 & .059 & .154 & .074 & .024 & .178 & .212 \\
                    \specialrule{0pt}{1pt}{1pt}
                    ManTraNet~\cite{Wu_2019_CVPR} & .349 & .385 & .375 & .398 & .360 & .334 & .244 & .333 & .096 & .105 & .266 & .247 \\
                    IML-ViT~\cite{ma2023iml} & .064 & .087 & .034 & .048 & .288 & .053 & .044 & .052 & .034 & .022 & .067 & .117 \\
                    \specialrule{0pt}{1pt}{1pt}
                    SAM~\cite{kirillov2023segment} & .117 & .133 & .054 & .087 & .141 & .213 & .053 & .124 & .184 & .149 & .069 & .033 \\
                    SAM-BBox~\cite{kirillov2023segment} & .061 & .055 & .029 & .041 & .058 & .076 & .034 & .092 & .099 & .062 & .018 & .017 \\
                    SAM2~\cite{ravi2024sam2} & .106 & .119 & .062 & .073 & .225 & .324 & .126 & .229 & .278 & .317 & .153 & .150 \\
                    SAM2-BBox~\cite{ravi2024sam2} & .042 & .038 & .015 & .022 & .044 & .076 & .035 & .085 & .088 & .061 & .016 & .018 \\
                    UniverSeg~\cite{butoi2023universeg} & .466 & .422 & .422 & .596 & .336 & .405 & .358 & .378 & .437 & .476 & .379 & .336 \\
                    SegGPT~\cite{wang2023seggpt} & .117 & .146 & .078 & .107 & .103 & .281 & .092 & .209 & .196 & .153 & .107 & .124 \\
                    \bottomrule[1pt]
                \end{tabular}
            }
        \end{minipage} \\
    \end{tabular}
    \label{tab:cos}
    \vspace{-15pt}
\end{table*}

\subsubsection{Concealed Object Segmentation}
\begin{figure}[t]
    \scriptsize
    \centering
    \begin{overpic}[width=.85\linewidth]{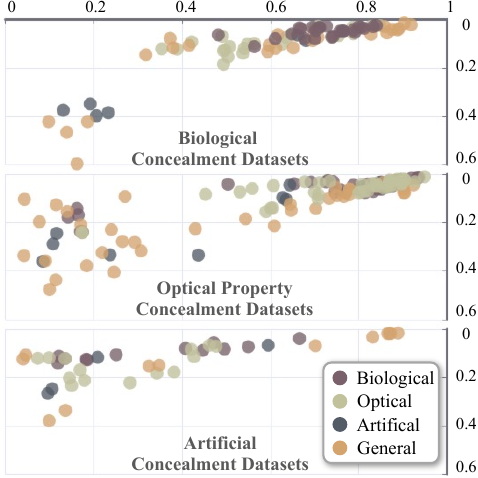}
        \put(44, 100){$wF_\beta$}
        \put(99, 48){$M$}
    \end{overpic}
    \caption{\small Concealment object segmentation capabilities across different datasets.
    }
    \label{fig:cos_vis}
    \vspace{-5pt}
\end{figure}

We selected several 19 representative methods to investigate recent advances in the concealed object segmentation (COS) and evaluated their generalizability across different concealed datasets.
%
As shown in \tabref{tab:cos}, datasets are grouped by concealment mechanisms.
%
For biological concealment, we followed Fan~\etal\cite{fan2020COD10K}, training on a mixed dataset of COD10K and CAMO. 
For optical concealment, the models were trained and test on individual datasets. 
%
For artificial concealment, we used Chen~\etal\cite{chen2021image}’s setup, training on CASIA-v2 and evaluating on two test sets. 
%
Results are visualized in \figref{fig:cos_vis}. 
%
%
Additionally, we included general methods not specific to any category to assess the feasibility of universal approaches to COS. Key findings include:

\noindent\textbf{$\bullet$}
Various COS tasks share a common objective: to recognize targets which concealed in the background. 
Consequently, the methods used across these subtasks show a high level of consistency in data preparation, model prediction, and result post processing, which suggests the potential for a unified model for different COS subtasks. 
Nevertheless, the subtle differences in concealment characteristics and solutions across different subtasks remain worthy of attention.

\noindent\textbf{$\bullet$}
The biological concealment counteraction approach demonstrates consistent performance across both biological and optical concealment perception subtasks, indicating that both tasks can benefit from a more robust encoder-decoder architecture. 
By enhancing the feature extraction capabilities of the encoder, the model can focus on both global and local information within the image, effectively identifying and localizing concealed targets by discerning the differences between foreground and background. 
The decoder aggregates information from multiple feature layers, facilitating the integration and prediction of results. 
FSPNet \cite{huang2023feature} achieves optimal performance in this subtask by designing encoders that utilize non-local token enhancement and a feature shrinkage decoder.

\noindent\textbf{$\bullet$}
However, each method experiences significant performance degradation when tasked with detecting artificial concealment targets. 
This decline can be attributed to two main reasons. 
First, artificial concealment fundamentally differs from the other two concealment mechanisms. 
On one hand, the concealment of the target occurs at the image editing level rather than in the physical world. 
On the other hand, the target's location is highly semantically plausible due to human involvement, rendering high-level semantic features nearly ineffective in recognizing them.
In this case, relying on low-level visual features, such as texture, would be more reliable.
Secondly, artificial concealment manifests in various forms, often exhibiting greater differences between mechanisms than those seen in biological c concealment. 
This variability complicates the task of learning to distinguish between multiple concealment modalities from a unified training set. 
Furthermore, this performance degradation is exacerbated by the significant domain gap between test sets and training sets.

\noindent\textbf{$\bullet$}
These methods from three concealment counteraction subtasks excel at detecting shadow regions in optical concealment. This is because shadows exhibit relatively consistent appearance patterns among all target objects or visual areas with concealment properties. The reason for this lies in the variations in light intensity caused by occlusions in the target scene, which are easily captured by models unless the lighting is dark or non-shadow areas have a dark appearance that weakens this variation. 

\noindent\textbf{$\bullet$}
Situations containing non-concealed targets are present in all three types of datasets. These images are used to enhance the model's understanding of concealment attributes. 
Training with non-concealed data improves the model's ability to distinguish between concealed and non-concealed objects, thereby reducing the likelihood of generating predictions for non-camouflaged targets. 
Notably, methods designed for biological concealment counteraction have made fewer attempts to utilize this data, typically focusing on concealed images. 
In contrast, techniques tailored for optical and artificial concealment counteraction often fully leverage this data by incorporating auxiliary networks or branches to differentiate concealment attributes.

\noindent\textbf{$\bullet$}
For uniform segmentation methods, SAM \cite{kirillov2023segment} is one of the key approaches that has garnered significant attention. 
The development of SAM is grounded in technological advances in natural language processing (NLP), and the introduction of prompt engineering has enabled it to demonstrate remarkable adaptability across various downstream tasks. 
The optimal performance exhibited by SAM here is attributed to its experimental setup. Specifically, to ensure that manual interactions do not affect the fairness of the evaluation, we refrain from inputting prompts. SAM focuses on every regions within the input image, identifies all objects, and generates segmentation results. Then we select the one with the highest Intersection over Union (IOU) with the ground truth for evaluation. 

Both UniversSeg \cite{butoi2023universeg} and SegGPT \cite{wang2023seggpt} are tailored for the uniform binary segmentation task,  establishing a new prediction paradigm for it. 
By providing a set of image-GT pairs during prediction, the model is directed towards the target to be segmented. 
However, neither method yields effective results when faced with concealed targets, which can be attributed to the inherent nature of these approaches. 
This paradigm emphasizes information based on semantic categories; specifically, it expects to identify targets in the input image that belong to the same category as those labeled in the reference image. 
This approach proves ineffective in the context of COS tasks. 
The model fails to concentrate on the concealment attribute, and the diversity of targets is so vast that it becomes challenging to provide enough cues from the reference image pair.

Additionally, the subpar performance of UniversSeg may stem from its design for medical image segmentation, which utilizes greyscale maps as model inputs. 
Medical images, primarily in the form of radiographs and other imaging modalities, are mostly single-channel greyscale images that lack the concept of depth of field. 
Furthermore, the use of a single light source and specialized detection methods results in noise patterns that differ significantly from those in natural images. 
%
Greyscale conversion causes major information loss, and domain gaps further harm model performance, leading to poor results.

\begin{table}[!t]
    \vspace{-15pt}
    \centering
    \scriptsize
    \renewcommand{\arraystretch}{1.2}
    \setlength{\tabcolsep}{3.8pt}
    \captionsetup{format=plain,labelformat=empty}
    \caption{
        \begin{tcolorbox}[colback=captionbg, colframe=white, sharp corners=south, boxrule=0mm, width=.99\linewidth]
            \small \textbf{TABLE 3:}\textbf{~Comparison of Concealed Object Detection.}                
        \end{tcolorbox}
    } 
    \vspace{-15pt}
    \label{tab:detection}
    \begin{tabular}{lcccccccc}
    \specialrule{1pt}{0pt}{0pt}
       \rowcolor{headerbg} Dataset & Method & Epoch & $AP$ & $AP_{50}$ & $AP_{75}$ & $AP_S$ & $AP_M$ & $AP_L$ \\
        \hline
        \rowcolor{tablebg} & DETR & 300 & 48.4 & 74.5 & 50.5 & 6.8 & 30.5 & 53.6 \\ 
        \rowcolor{tablebg}& HTC & 12 & 27.9 & 56.9 & 23.8 & 5.6 & 15.7 & 31.4 \\ 
        \rowcolor{tablebg}& DINO & 12 & \textbf{53.8} & \textbf{78.4} & \textbf{56.5} & \textbf{12.8} & \textbf{38.0} & \textbf{58.4} \\ 
        \rowcolor{tablebg}\multirow{-4}{*}{Trans10K} & YOLOv11 & 200 & 28.2 & 55.6 & 25.1 & 4.9 & 13.8 & 32.1 \\  \specialrule{0pt}{0.5pt}{0.5pt}
        \rowcolor{tablebg}& DETR & 300 & 81.4 & 89.8 & 83.6 & 6.2 & 11.4 & 83.1 \\  
       \rowcolor{tablebg} & HTC & 12 & 70.9 & 88.1 & 75.8 & 0.1 & 6.1 & 72.3 \\ 
       \rowcolor{tablebg} & DINO & 12 & \textbf{84.3} & \textbf{91.6} & \textbf{86.1} & \textbf{8.1} & \textbf{16.7} & \textbf{85.8} \\ 
       \rowcolor{tablebg} \multirow{-4}{*}{COD10K} & YOLOv11 & 200 & 81.5 & 90.2 & 83.3 & 6.7 & 15.3 & 83.1 \\ \specialrule{0pt}{0.5pt}{0.5pt}
       \rowcolor{tablebg} & DETR & 300 & 54.5 & 73.1 & 57.8 & 8.0 & 46.1 & 71.9 \\ 
       \rowcolor{tablebg} & HTC & 12 & 50.5 & 67.9 & 55.1 & 6.1 & 42.7 & 65.6 \\ 
       \rowcolor{tablebg} & DINO & 12 & \textbf{61.0} & \textbf{77.2} & \textbf{63.4} & \textbf{13.1} & \textbf{55.2} & \textbf{76.7} \\ 
       \rowcolor{tablebg} \multirow{-4}{*}{PMD} & YOLOv11 & 200 & 55.9 & 70.2 & 58.5 & 9.1 & 48.7 & 71.9 \\ \specialrule{0pt}{0.5pt}{0.5pt}
       \rowcolor{tablebg} & DETR & 300 & 34.6 & 53.8 & 34.6 & 12.0 & 39.5 & 64.1 \\ 
       \rowcolor{tablebg} & HTC & 12 & 45.0 & 65.1 & 47.5 & 25.1 & 50.0 & 70.0 \\ 
       \rowcolor{tablebg} & DINO & 12 & \textbf{48.3} & \textbf{65.9} & \textbf{50.9} & \textbf{30.4} & \textbf{53.4} & \textbf{72.3} \\ 
       \rowcolor{tablebg} \multirow{-4}{*}{SBU} & YOLOv11 & 200 & 46.2 & 64.4 & 49.0 & 28.2 & 50.5 & 72.3 \\ \specialrule{0pt}{0.5pt}{0.5pt}
       \rowcolor{tablebg} & DETR & 300 & 38.5 & 50.7 & 41.0 & 17.0 & 37.8 & 53.0 \\ 
       \rowcolor{tablebg} & HTC & 12 & 32.2 & 43.2 & 34.8 & 16.6 & 33.8 & 38.5 \\ 
       \rowcolor{tablebg} & DINO & 12 & \textbf{47.2} & \textbf{57.4} & \textbf{49.8} & \textbf{21.8} & \textbf{46.4} & \textbf{64.7} \\ 
       \rowcolor{tablebg} \multirow{-4}{*}{CASIA} & YOLOv11 & 200 & 39.7 & 50.4 & 41.7 & 20.1 & 38.5 & 56.8 \\
        \specialrule{0.8pt}{0pt}{0pt}
    \end{tabular}
\end{table}

\subsubsection{Concealed Object Detection}
To further explore the impact of concealment mechanisms on dense prediction, we quantitatively evaluated the concealed object detection task.
We selected five concealed object segmentation datasets featuring representative concealment mechanisms and transformed them to match the object detection annotation format in COD10K. 
This process resulted in a simple concealed object detection benchmark containing a single class (\ie, the foreground class), allowing us to evaluate the deep model's ability to detect concealed foregrounds against the full image.
We present the quantitative evaluation results of four common object detection methods in \tabref{tab:detection}, highlighting the following key findings:

\noindent\textbf{$\bullet$}
In quantitative experiments involving four classical methods, DINO achieved optimal performance across five datasets. 
This improvement can be attributed to its contrastive denoising strategy, which enhances the model's ability to differentiate between positive and negative samples. It is crucial for effectively distinguishing between the foreground and background in concealed object detection tasks.
Additionally, YOLOv11, an advanced real-time object detection model, achieved notable results. This suggests that optimizations in network structure and improvements in feature extraction, driven by advancements in deep learning technology, enhance the detection of concealed targets.
\noindent\textbf{$\bullet$}
None of these methods achieve satisfactory performance on the concealed dataset. Even when the foreground object is simplified to a single class, none exceed 90\% in AP metrics, with most hovering around 50\%. This indicates that the negative impact of the concealment mechanism on visual perception also affects object detection.
Furthermore, the model's predictions yield significantly more false negatives than false positives, with a ratio of 5:1. This suggests that concealment primarily affects the model by causing it to overlook the target. The few false positive results primarily arise from the complex appearance and presentation of concealed targets, which lead the model to mistakenly identify small portions of the target as separate concealed targets.

\noindent\textbf{$\bullet$}
The same methods show large differences across different datasets.
All methods achieved favorable results on the Trans10K dataset, primarily due to its focus on objects like glass cups and display windows, which feature a limited number of categories and a consistent appearance. 
In contrast, the results on the other datasets demonstrate varying degrees of difficulty. 
The COD10K dataset contains 78 categories, and the complexity of the foreground classes may hinder the model's ability to capture a unified concealment characteristic, resulting in degraded performance. 
The targets of remaining three datasets predominantly consist of regions rather than distinct objects, a challenge particularly evident in the CASIA and SBU, where detection models struggle to recognize the complex appearances and varying shapes of different regions.
Additionally, these methods are affected by the intricate indoor environments in the PMD and the numerous small shadow areas in the SBU dataset.

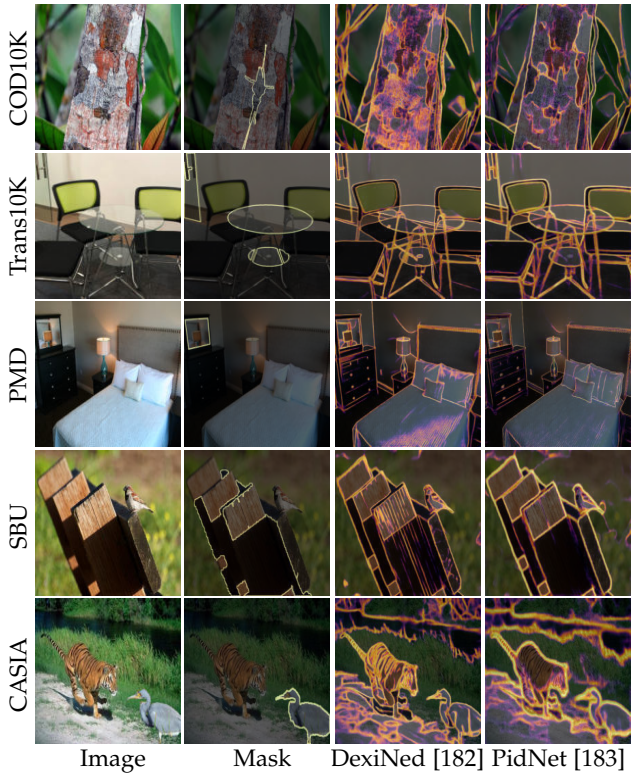
\begin{figure}[!t]
    \centering
    \begin{overpic}[width=.95\linewidth]{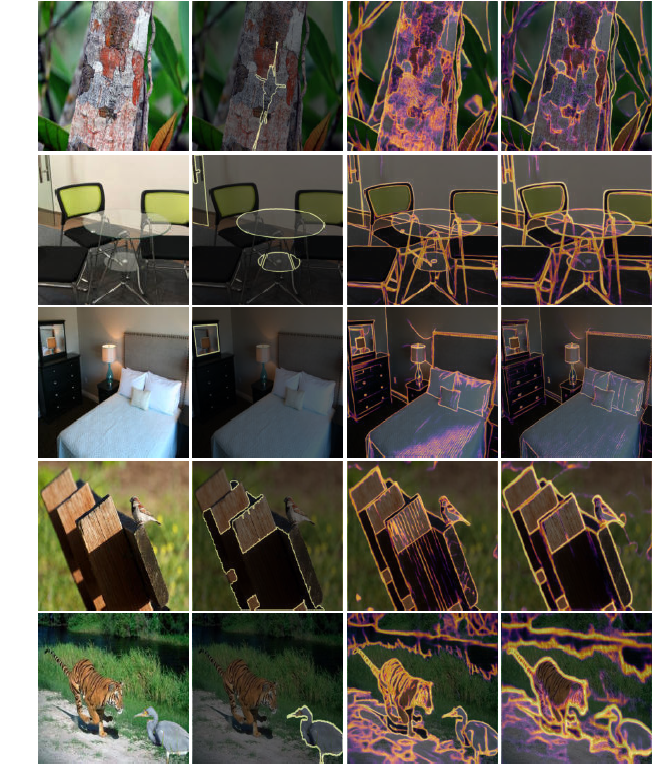}
        \put(1.6, 5.2){\small\rotatebox{90}{CASIA}}
        \put(1.6, 27){\small\rotatebox{90}{SBU}}
        \put(1.6, 46.5){\small\rotatebox{90}{PMD}}
        \put(1.6, 63.5){\small\rotatebox{90}{Trans10K}}
        \put(1.6, 83.5){\small\rotatebox{90}{COD10K}}
        \put(11, -2){\small{Image}}
        \put(31.5, -2){\small{Mask}}
        \put(44.5, -2){\small{DexiNed\cite{xsoria2020dexined}}}
        \put(66, -2){\small{PidNet\cite{pdc-PAMI2023}}}
    \end{overpic}
    \vspace{4pt}
    \caption{\small Visualization of concealed edge estimation. 
        } 
    \label{fig:edge_detection}
\end{figure}

\subsubsection{Concealed Edge Estimation}
We conducted a quantitative evaluation using two general edge estimation deep models on the concealed datasets, with \figref{fig:edge_detection} visualizing several representative examples.
The results primarily exhibit two types of performance: First, in the COD10K dataset, animals use background matching as the main concealment strategy, which poses challenges for edge estimation. 
On the one hand, the targets have indistinct boundaries, and their highly integrated visual features with the background make it difficult for models to accurately capture the target edges, often resulting in predictions that lack the true boundaries. 
On the other hand, the edges of the targets do not stand out against the cluttered background. These phenomena demonstrate the importance of edges for the visual perception of concealed targets and provide insight into why the edge cues based method introduced in \secref{sec:Additional_Cues} can enhance model performance.

In contrast, concealed targets in the other datasets typically have distinct boundaries, leading to predictions that usually include their true edges. 
These targets often employ ``misleading'' rather than ``disappearing'' concealment strategies; even when fully visible, they may be misclassified by the visual system as belonging to other categories. 
Also, they are still affected by cluttered backgrounds, making it challenging to separate them from their surroundings.

\section{Discussion and Outlook}

\subsection{Applications}

\noindent\textbf{$\bullet$ Industry.}
Vision technology in industry detects surface defects caused by physical factors like friction and impact, or chemical factors such as corrosion, which manifest as scratches, color disparities, fractures, and stains. 
Despite their subtle appearance, these defects pose significant threats to industrial production processes.
To address this challenge, Fan~\etal\cite{fan2023advances} collected and analyzed a concealed defect segmentation dataset, CDS2K, to assess the applicability of CDP methods in industrial scenarios. 

Further, when concealed defects appear on the facade of a building, they threaten inhabitants' safety. 
Although techniques such as laser scanning or SLAM are replacing visual inspection, they are still limited by factors such as time and cost.
On the contrary, vision-based inspection of facade defects can enhance efficiency, reduce costs, and have great practical application prospects.

\noindent\textbf{$\bullet$ Agriculture.}
%
In high-density planting environments, small targets and dense, overlapping occlusion between plants pose significant challenges for accurate crop detection.
Wang~\etal\cite{wang2024depth} introduced a new dataset, ACOD-12K, for multimodal crop detection with depth cues, marking a significant step in concealed crop detection.

\noindent\textbf{$\bullet$ Medicine.}
In medicine, lesion areas often exhibit distinct visual patterns or anomalies, making images crucial aids in diagnosis.
Using computer technology to automatically analyze pathological images and provide recommendations can streamline the diagnostic process and reduce the workload of healthcare professionals.
However, challenges persist, such as in polyp segmentation~\cite{ji2024frontiers}, where issues such as variations in color distribution within the lesion area, small target sizes, and indistinct boundaries between lesions and normal tissue arise.
Moreover, noise introduced during data acquisition in the complex environment of the human body, the incorporation of spatial information from specialized data sources, and the complexity of target features further compound the difficulty in identifying such targets. 

\noindent\textbf{$\bullet$ Autonomous Driving.}
In autonomous driving, the focus is on vehicle perception, decision-making, and maneuvering. 
Vision algorithms, along with LiDAR, are crucial for perception. 
By emulating human vision using cameras and employing deep neural networks for classification, detection, and segmentation, self-driving vehicles gather detailed environmental data. 
Although cost-effective and convenient, cameras cannot replicate the human eye's three-dimensional perception and may suffer distortions in complex environments, such as inclement weather or low visibility, potentially leading to target perception failures and severe accidents. 
Addressing concealment caused by adverse weather conditions is a key concern in this application.
%

Additionally, CDP is crucial in wildlife conservation, human rescue operations, and public safety assurance \cite{xu2023deepchange}.
%
These scenarios highlight the diverse potential of CDP across various fields, demonstrating its adaptability and value in addressing different domain-specific challenges.
%
%
Exploring CDP is expected to yield significant benefits for multiple industries and disciplines in the future.

\subsection{Towards the Unified Approach} 
\label{sec:cvpagent}
While researchers have made notable efforts across various domains, existing studies on concealed vision remain largely constrained by specific target categories, application scenarios, and task types, highlighting a considerable gap toward achieving Artificial General Intelligence (AGI).
As more complex concealment mechanisms and data modalities are gradually being mined and defined, repeating the past simple research approach of constructing new data and migrating model architectures to new tasks is labor-intensive and inefficient.
Recent advancements in multimodal large language models (MLLMs) have demonstrated exceptional capabilities in perception and reasoning. 
By understanding user intentions, decomposing intricate tasks, and dynamically executing instructions, these models facilitate efficient collaboration between large and small models, making it possible to construct a unified agent for concealed vision tasks.
In this context, we restructured \data, a multimodal instruction-tuning dataset tailored to the concealed visual perception domain. 
%
Using this dataset, we have developed a unified intelligent agent, \agent, which can fully unleash the powerful semantic understanding capabilities of LLMs in the concealed vision domain through effective human-computer interaction.
\subsubsection{Instruction Tunning Dataset: CvpINST}
\begin{figure*}[t]
    \vspace{10pt}
    \centering
    \begin{overpic}[width=1\linewidth]{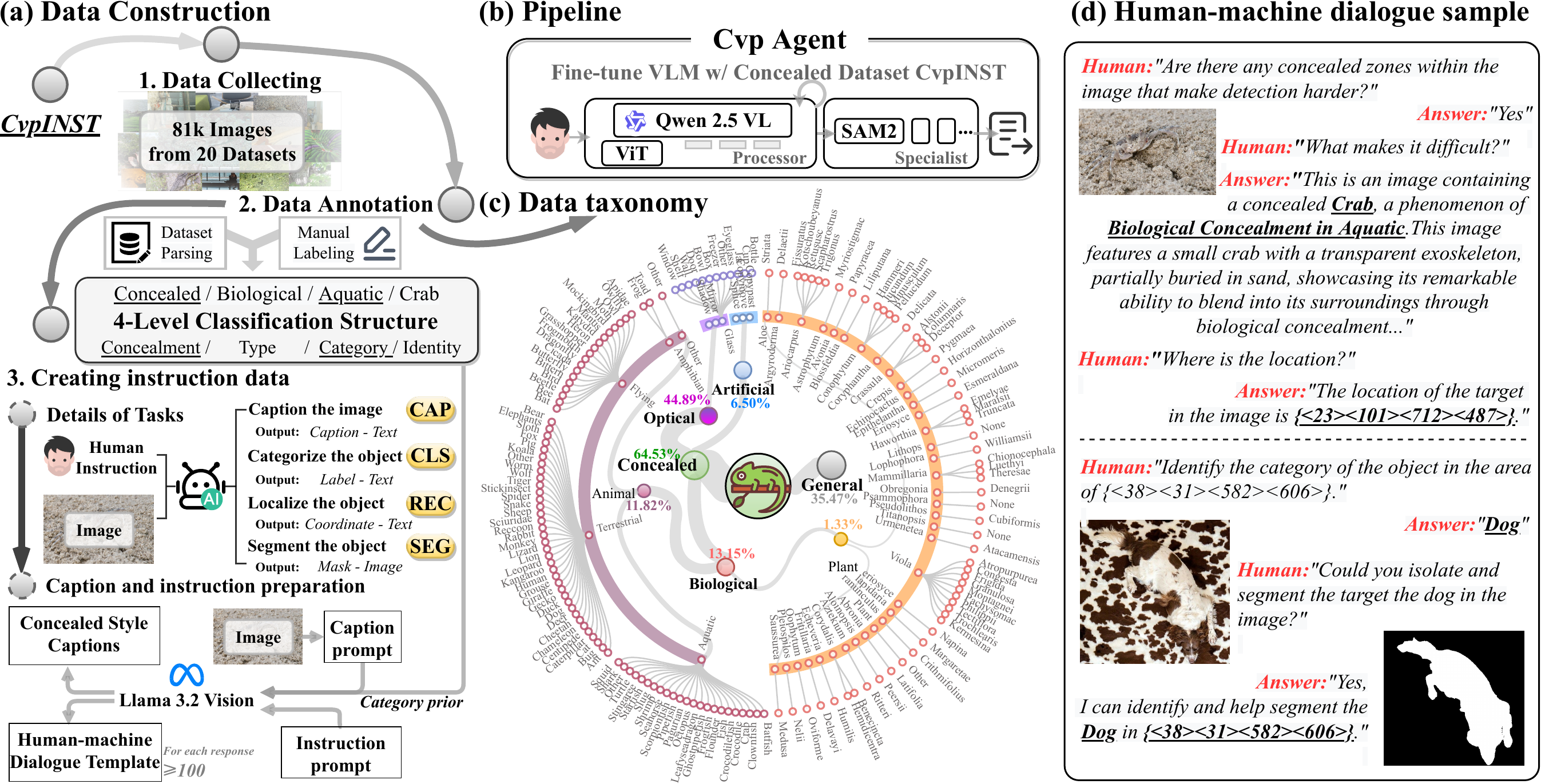}
    \end{overpic}
    \caption{\small Details of \data ~and \agent.
    }
    \label{fig:cvpagent}
\end{figure*}

\noindent\textbf{$\bullet$ Dataset collection.}
\data ~consists of 81,279 images, including 47,709 concealed images and 33,570 general images, collected from 20 publicly available datasets. We aggregated 13 widely studied datasets in the concealed vision domain as the source of concealed data and adhered strictly to the original train-test splits to ensure no data leakage.
The concealed data covers all concealment types discussed in this review. Original category labels and binary masks indicating target locations were retained in the dataset. To enhance generalizability and avoid potential overfitting, we collected non-concealed images and their binary masks from seven commonly used saliency detection datasets. These images were randomly shuffled and divided into training and test sets in a ratio of 75\% / 25\%.

\noindent\textbf{$\bullet$ Data labeling.}
We organized the images using a partially hierarchical four-level classification structure to inject the structured classification theory on concealment taxonomy proposed in this review into the dataset. As shown in \figref{fig:cvpagent}(a)-2, the images are categorized based on whether they are concealed, concealment type, target category, and target name, forming a four-level hierarchy with 2 roots / 3 parents / 36 children / 128 leaves.
For non-concealed images, specific categories are not considered; they are uniformly assigned to a ‘General’ root class without further subdivision. For concealed images, datasets such as COD10K \cite{fan2020COD10K}, PlantCamo \cite{plantcamo}, and Trans10K \cite{xie2020trans10k} already include classification labels, which are naturally mapped to the child and leaf levels. These images were also annotated with concealment-type labels as parents based on the classification criteria defined in this review. For example, a concealed crab would be classified as Concealed/Biological/Aquatic/Crab.
To maintain consistency, we additionally annotated the target names for 8,544 images from other biological concealment categories and the glass category under optical concealment. For other concealed target types, such as shadows and mirrors, which are not associated with specific target names, only three levels of classification are provided, e.g., Concealed/Optical/Mirror.
The hierarchical category structure of the \data~ dataset is visualized in \figref{fig:cvpagent}(c).

To further explore the underlying causes of concealment phenomena and provide reasonable explanations for the concealment processes of targets in images, we enhanced the dataset by adding explanatory captions. Taking into account the distinctive characteristics of different concealment mechanisms, we introduce a template-based pipeline to customize caption templates for various types of concealment. As \figref{fig:cvpagent}(a)-3 depicts, through interactions with Llama 3.2 Vision, we generated detailed descriptions for every image.
Specifically, the hierarchical category structure of the target image was incorporated into the template, leveraging category priors to enable the multimodal large language model (MLLM) to comprehend the image better. In addition, the pipeline included a concise summary of the characteristics of different types of concealment. For example, 'Optical concealment means that the target is not easily detectable due to its optical properties, in this case, the transparent nature of glass'. This helped the model infer the reasons for concealment based on existing knowledge.
Finally, our captions placed greater emphasis on the relationship between the concealed target and its environment, and the factors contributing to the concealed effect.

\noindent\textbf{$\bullet$ Instruction preparation.}
Finally, we construct one-round or multi-round conversational dialogue formats using organized image-text pairs and category labels. 
As illustrated in \figref{fig:cvpagent}(a)-3, we employed Llama to generate dialogue templates encompassing four distinct task types, following the approach described in \cite{ji2024frontiers}.
Specifically, the caption task (CAP) is designed to comprehensively analyze whether an image contains concealment, the category of concealment, the hierarchical category, and the reasons behind the concealment. The classification task (CLS) requires the model to output the category of a concealed target at a specified location. The recognition task (REC) aims to output the coordinates of targets belonging to a given category. The final segmentation task (SEG) is intended to output the mask of a target at a specified location. 
To ensure data diversity, we generated over 100 templates for each response in the multi-round dialogues. When generating data for these four tasks for each sample, dialogues are randomly selected from the templates. For a 2-round human-machine dialogue, this theoretically results in $ 100^4 $ possible variations, thereby preventing the model from overfitting to specific expressions.

\subsubsection{CvpAgent for Concealed Visual Perception}
Our \texttt{CvpAgent} has two components: a processor built upon a MLLM and expert tools equipped for different downstream tasks. 
Specifically, we fine-tuned the latest Qwen-VL model
\cite{Qwen2.5-VL} 
using LoRA \cite{hu2022lora} and our \texttt{CvpINST} dataset. 
As illustrated in \figref{fig:cvpagent}(a)-3, the model accepts images and human instructions, subsequently generating corresponding outputs. 
For the first three tasks, the model only needs to produce textual outputs, which can be accomplished solely by the MLLM. 
The fourth task requires the model to excel in the segmentation domain and possess image output capabilities. 
To ensure the scalability of the proposed \agent, we designed the instruction data following the approach of Vitron \cite{fei2024vitron}. 
For complex tasks, the MLLM is responsible for dialogue and intent recognition. Once the user’s intent is determined, it outputs a model label indicating the need to invoke a specific expert model, while integrating existing clues as inputs to the expert model. 
We integrated SAM2 \cite{ravi2024sam2} as the expert model for the segmentation task, where the MLLM performs task judgment and extracts target category and location information as inputs. 
%
Consequently, we have established a unified and extensible framework for camouflage visual perception, which allows for the flexible replacement of the MLLM and downstream expert models to better adapt to various real-world application scenarios.

\subsubsection{Broader Implications}
\noindent\textbf{$\bullet$ Data diversity.}
CDP tasks often rely on single-data sources, such as focusing on animal or industrial defects. We propose a cross-modal, cross-domain dataset solution that is not restricted to specific target categories, enabling broader applicability in real-world scenarios.
%

\noindent\textbf{$\bullet$ Generalized applicability.}
Expanding the boundaries of concealed vision, we extend beyond standalone segmentation tasks to encompass multi-task joint modeling, including recognition, detection, and segmentation. This shift advances concealed vision research from the task level to the scene level, enhancing applicability in diverse and complex scenarios through human-computer interaction and elevating the overall level of intelligence.
%

\noindent\textbf{$\bullet$ Theoretical integration.}
By integrating the novel findings and the structured classification theory on concealment taxonomy proposed in this review into the dataset, \agent achieves a more nuanced and multidimensional understanding of concealed attributes. This allows the model to analyze the concealment mechanisms hierarchically, promoting the standardization of the definition and classification of concealed targets.
%

\noindent\textbf{$\bullet$ Interpretability.}
The unified paradigm of multi-task integration enables the model to identify concealed targets during interactions while providing the rationale behind its judgments. This transparent decision-making process ensures the interpretability of the results.

\subsection{Potential Research Direction}
%
\subsubsection{Data Acquisition and Annotation}
In recent years, deep learning methods have achieved state-of-the-art performance in many tasks within the field of CDP, aided by the collection and annotation of numerous high-quality datasets.
Despite efforts to enhance datasets by increasing data scale or annotations, data development still lags behind deep learning advancements in CDP, becoming a primary bottleneck limiting current deep learning methods' performance.
Thus, several open issues in data acquisition and annotation require further research, including data categories, annotation granularity, and dataset modality.
\noindent\textbf{$\bullet$ Category diversity.}
Currently, the diversity of data categories in the field of CDP is limited.
On one hand, research has mainly focused on a few target categories, such as animals, transparent objects, and mirrors, with little mention of concealed objects in open environments (such as concealed plants, common objects in living environments, or concealed defects on industrial product surfaces).
This limits researchers' exploration of a wider range of concealment properties and confines the application of CDP to specific scenarios \cite{kejriwal2024challenges}.
On the other hand, existing datasets generally only contain a single concealment category, lacking a comprehensive benchmark for researchers to study the relationship between different categories of concealment and further explore unified methods for CDP. A large-scale dataset that combines different concealment categories may be key to addressing this issue.
\noindent\textbf{$\bullet$ Fine-grained annotations.}
As shown in \tabref{tab:dataset}, datasets provide more than one type of annotation to meet different task requirements. High-quality annotations serve as the foundation for the effectiveness of deep learning methods \cite{day2023improving}.
Some datasets even offer multiple levels of fine-grained annotations, such as the structured annotations covering four levels provided by COD10K \cite{fan2020COD10K}, which have been further utilized in concealment ranking \cite{lv2023towards}, weakly supervised scribbles \cite{he2023weakly}, and fixation \cite{lv2021simultaneously}, among other works.
However, more detailed annotations are needed, especially concerning concealment attributes.
Providing annotations for concealed attributes can help researchers tailor network structures to address different concealment patterns.
%
COD10K represents a pioneering effort by describing high-difficulty attributes in natural scenes with six attributes.
COD-TAX \cite{zhang2024unlocking} further investigates the contribution of concealed attributes to concealed scenarios by providing a detailed description from three main categories and 17 attribute-level categories.

\noindent\textbf{$\bullet$ Multimodal data.}
Existing datasets have been enriched with additional visual cues for more comprehensive annotations, but this enrichment is limited to the visual modality.
Pang~\etal\cite{pang2024open} made significant strides by incorporating text information into the concealed target segmentation task.
Multimodal data helps models perceive concealed targets from different dimensions.
Leveraging multimodal datasets enables researchers to explore the complex interplay between visual, auditory, and other sensory cues in concealment, leading to advancements in both theoretical understanding and practical applications across various fields.

\subsubsection{Unified Concealed Perception Scheme}
The binary segmentation task, which focuses on pinpointing and precisely defining the boundary between target and background regions,  has been widely studied in CDP as a basic 
concealment counteraction task.
Typically, existing studies concentrate on specific concealment schemes, tailoring datasets and model architectures accordingly to address individual concealment mechanisms.
However, due to the highly diversified mechanisms of concealment, developing separate models for each concealment scenario is inefficient. Practical application demands solutions capable of accommodating model generalization.
%

Recent efforts have explored the development of a unified pipeline capable of handling targets with diverse concealments.
Lv~\etal\cite{lv2021simultaneously} introduced a model encompassing localization, segmentation, and sorting objectives, marking an initial step in this direction. 
Building upon this, Liu~\etal\cite{liu2023explicit} drew inspiration from the unified NLP foundation model and proposed a novel approach that enables a visual prompting model to address four distinct segmentation tasks simultaneously.
Zhao~\etal\cite{zhao2024spider} extended this approach by unifying eight segmentation tasks using context-dependent concepts, arguing that these tasks emphasize conceptual understanding and have functional applications broader than traditional semantic segmentation.

Additionally, there is potential for developing generic models capable of handling multimodal inputs and various tasks.
The large-scale multimodal instruction fine-tuning dataset \data~ and the unified concealed vision perception \agent~ constructed in \secref{sec:cvpagent} are a concrete attempt at this idea in action.
By introducing the SOTA MLLM as the core, these models are able to learn the prior knowledge in the concealed vision community, and then accomplish complex tasks through efficient collaboration among models.
%
This offers insights into the future of concealed vision while promoting CDP’s practical applications and progress toward general artificial intelligence.

\subsubsection{Data-Efficient CDP Model}
%
Diverse, large concealed datasets are pivotal for advancing CDP. 
However, owing to the unique nature of concealment, both data collection and annotation incur significantly higher costs compared to other domains. 
An instance-level segmentation annotation for a single image can take up to 60 minutes \cite{fan2021concealed}, making it challenging to scale dataset sizes or obtain data from videos and other modalities.
Therefore, the adoption of data-efficient methods for training deep models in CDP is crucial.
Current research focuses primarily on weakly supervised learning, where sparse annotations are typically derived from manually labeled scribbles \cite{he2023weakly,niu2024minet}, points \cite{chen2024just}, boxes \cite{zhang2024learning} or pre-segmentation results of other segmentation models \cite{he2024weakly,chen2024sam,he2024text}. 
These approaches mitigate the bottleneck caused by data annotation to some extent.
In addition, annotation difficulties may compromise label reliability and completeness, highlighting the importance of addressing the effects of noisy and structure-missing data on model training \cite{fu2024semi, mitra2023learning}.
%
Regarding other schemes such as self-supervised/semi-supervised learning \cite{lai2024camoteacher}, few-shot learning, and transfer learning, related research in this community is still in its early stages and calls for further exploration.

\subsubsection{Concealment Generation and Assessment}
Recently, studies have revealed the potential of using synthetic data to enhance existing tasks \cite{liang2022advances,pati2024accelerating}. Understanding the generation process of concealment and creating concealed samples can help uncover its underlying mechanisms and impacts.
%
The task of camouflage image generation was introduced in 2010 \cite{chu2010camouflage}, with a large number of recent efforts \cite{zhang2020deep,li2022location} exploring methods to create concealed images customized to specific objects. 
However, due to the absence of systematic research on concealment mechanisms, style transfer-based generation results have yet to find practical applications.
At the application level, Zhao~\etal\cite{zhao2024camouflaged}. achieved automatic large-scale concealed image generation by a new inpainting-based framework, which provides the possibility of expanding the diversity of concealed object categories.
Some studies 
\cite{lamdouar2023making,zhang2023camouflaged} 
have also utilized camouflage image generation as a data augmentation technique, generating high-quality images to enrich datasets and enhance model training accuracy.
As the study of concealment progresses, deeper insights into concealment mechanisms will propel the rapid development of camouflage generation tasks.
Furthermore, the field of CDP has long lacked standardized assessment metrics due to the intricate relationship between concealment and human perception. 
Nevertheless, developing concealment evaluation metrics holds significant importance for assessing generation quality and refining fine-grained perception methods.
In this context, Lamdouar~\etal\cite{lamdouar2023making} pioneered a composite index that incorporates reconstruction difficulty, boundary visibility, and distribution disparity between target and background, laying the foundation for an effective concealment assessment.
Looking ahead, the construction of a comprehensive concealment assessment method encompassing various concealment types is poised to become a focal point of research in the CDP community.

\subsubsection{Combining Large Models}
OpenAI o1 \cite{openai2024learning} and DeepSeek-R1 \cite{deepseekai2025deepseekr1incentivizingreasoningcapability} have opened up a new era of large reasoning models.
Thanks to the multimodal reasoning ability and excellent generalization of large models, recent research has demonstrated that combining them with task-specific methods often improve performance.
%
In particular, when addressing multimodal tasks, large models can replicate how humans naturally process multi-sensory inputs and collaborate in a complementary manner, leading to enhanced reasoning capabilities.
%
Techniques that combine large models include downstream task fine-tuning, model distillation, model compression, and fixed parameters for feature extraction.

There have been some efforts in the CDP community to integrate large models.
An intuitive approach involves employing a pre-trained large foundation model as a feature extractor and then fine-tuning it on a CDP task \cite{xing2023pre,luo2023vscode,yu2024exploring,zhang2024COMPrompter}. 
This process facilitates the transfer of comprehensive knowledge acquired from diverse multimodal datasets to the specific CDP task.
Moreover, these foundation models can serve as knowledge repositories, offering multimodal understanding prior \cite{cheng2023large} or generalized prompts \cite{hu2023relax}.
These large models' robust comprehension, reasoning, and multimodal processing capabilities offer avenues for the CDP community to progress toward an open environment. 
This also establishes a technical groundwork for the emergence of a more unified, generalized, and multimodal approach to the CDP area.
\subsubsection{Novel Tasks and Advanced Paradigms}
Deep learning is advancing rapidly. On one hand, as research in concealment vision progresses, a series of new tasks emerge. From images~\cite{fan2020COD10K} to videos~\cite{cheng2022implicit}, from  segmentation~\cite{fan2021concealed} to ranking~\cite{lv2021simultaneously} and counting~\cite{sun2023iocformer}, and from single domains to related fields~\cite{fan2023advances}, the introduction of new tasks often injects fresh vitality into the field. In addition to the tasks mentioned above, recently, Zhang~\etal\cite{zhang2023referring} and Cheng~\etal\cite{cheng2023large} have studied the Ref-COD task, which involves segmenting specified concealed targets based on a small group of reference texts or images containing salient objects. The emergence of the open-vocabulary camouflaged object segmentation (OVCOS) task~\cite{pang2024open,vu2023leveraging} explores the path for CDP leveraging large language models to navigate an open world. Furthermore, to address the lack of data in the CDP community, researchers have also shown interest in zero-shot~\cite{li2023zero,tang2024chain} and few-shot~\cite{nguyen2023art} learning.
On the other hand, advanced network architectures and learning paradigms can also enhance existing tasks. Mao~\etal\cite{mao2021generative} introduced GANs into concealed target detection to avoid the overconfidence issues faced by CNN and Transformer frameworks. 
Some methods \cite{chen2024camodiffusion,zhao2024focusdiffuser,sun2025conditional} integrated the progressive refinement and stepped sampling processes of the Diffusion model into the camouflaged object detection task to address problems related to overconfidence and rough boundaries. 
Additionally, Zhao~\etal\cite{zhao2024towards} explored the mechanism of information exchange between the encoder and decoder in the traditional U-shaped structure, introducing a simple gating network to control information flow and establish synergy among information. Further investigation into the relationship between new network architectures, deep learning paradigms, and CDP may offer insights into addressing specific challenges within the community.

\section{Conclusion}
We review recent advances in artificial intelligence research in the CDP community.
By tracing the development of CDP and analyzing the causes and characteristics of concealment, we propose a hierarchical taxonomy that categorizes concealment. It helps to clarify the unique challenges of concealment and its impact on visual perception.
From the perspective of concealment counteracting, we conduct a review of deep learning methods for CDP tasks, aiming to provide researchers with a broad and structured understanding of current techniques.
In addition, we discuss application scenarios and outline future research directions.

Drawing from current trends and insights, key areas for future development in CDP over the coming years include: large-scale, high-quality datasets with richer categories, multiple modalities, and fine-grained annotations; unified frameworks capable of handling diverse concealment scenarios; data-efficient learning methods under weak, self-, or unsupervised supervision; generation of concealed samples and assessment of them with the human preferences; CDP methods integrated with paradigms such as reinforcement learning, adversarial learning, and continual learning; and the integration of large models to enhance reasoning and generalization capabilities.
To support this vision, we present a preliminary implementation by constructing a large-scale multimodal instruction fine-tuning dataset \data~ and a concealed visual perception agent \agent~, offering an initial exploration into the role of foundation models in CDP. Our goal is to inspire further interest and development in this area.

As these directions advance, CDP is expected to demonstrate increasing practical value. 
Enhanced CDP methods will benefit diverse applications, including industrial inspection, healthcare, crop yield estimation, pest control, marine species conservation, and intelligent security, addressing current systems’ limitations in detecting concealed targets.
%
This review aims to provide new researchers and practitioners with a comprehensive overview of the CDP field, facilitating future research and contributing to ongoing advancements in computer vision.

\par

\ifCLASSOPTIONcaptionsoff
  \newpage
\fi

\bibliographystyle{IEEEtran}
\bibliography{COD}





\end{document}